\documentclass{article}

\usepackage[preprint]{neurips_2026}

\usepackage{amsmath,amsfonts,bm}

\usepackage{amsmath,amsfonts,bm}

\newcommand{\eg}{\textit{e.g.}, }

\def\eqref#1{equation~\ref{#1}}

\def\1{\bm{1}}

\DeclareMathAlphabet{\mathsfit}{\encodingdefault}{\sfdefault}{m}{sl}
\SetMathAlphabet{\mathsfit}{bold}{\encodingdefault}{\sfdefault}{bx}{n}

\usepackage{adjustbox}
\usepackage[table]{xcolor}
\usepackage[utf8]{inputenc} 
\usepackage[T1]{fontenc}    
\usepackage[colorlinks,
            linkcolor=red,
            anchorcolor=green,
            citecolor=blue
            ]{hyperref}
\usepackage{url}            
\usepackage{textcomp}      
\usepackage{booktabs}       
\usepackage{amsfonts}       
\usepackage{nicefrac}       
\usepackage{microtype}      
\usepackage{caption}
\usepackage{tabularx}
\usepackage{siunitx}
\usepackage{multirow}
\usepackage{booktabs}
\usepackage{microtype}
\usepackage{graphicx}
\usepackage{subfigure}
\usepackage{tcolorbox}
\usepackage{makecell}
\usepackage{enumitem}

\usepackage{url}
\usepackage{flushend}
\usepackage{wrapfig}
\usepackage{booktabs} 
\usepackage{placeins}

\usepackage{algorithm}
\usepackage{algorithmic}

\usepackage{amsmath}
\usepackage{amssymb}
\usepackage{mathtools}
\usepackage{amsthm}

\definecolor{myred}{rgb}{0.8, 0.2, 0.2} 
\definecolor{myblue}{rgb}{0.2, 0.2, 0.8} 
\definecolor{TableTitleBg}{RGB}{63,63,63}
\definecolor{LightGrayBg}{RGB}{242,242,242}
\definecolor{LLMHeaderBg}{RGB}{220,220,220} 
\definecolor{WhiteText}{RGB}{255,255,255}
\definecolor{BlackText}{RGB}{0,0,0}
\definecolor{RedText}{RGB}{255,0,0} 

\usepackage[capitalize,noabbrev]{cleveref}

\title{UniVLR: Unifying Text and Vision in \\ Visual Latent Reasoning for Multimodal LLMs}

\author{%
    Houcheng Jiang$^{1,4}$\thanks{Equal contribution},
 Jiajun Fu$^{1*}$,  Junfeng Fang$^{3}$,  Chen Gao$^{2,4}$\thanks{Corresponding author: \texttt{\{chgao96, xiangwang1123\}@gmail.com}}, \\  \textbf{Xiang Wang$^{1\dagger}$,  Xiangnan He$^{1}$,  Yong Li$^{2,4}$} \\ $^1$University of Science and Technology of China,  $^2$Tsinghua University,\\  $^3$National University of Singapore, $^4$Zhongguancun Academy\\
    \texttt{\{janghc, fujiajun\}@mail.ustc.edu.cn}
}

\begin{document}

\maketitle

\begin{abstract}
Multimodal large language models are increasingly expected to perform \emph{thinking with images}, yet existing visual latent reasoning methods still rely on explicit textual chain-of-thought interleaved with visual latent tokens. 
This interleaved design limits efficiency and keeps reasoning fragmented across separate text and vision channels. 
We propose \textbf{UniVLR}, a unified visual latent reasoning framework that treats textual reasoning and auxiliary visual evidence as a shared visual workspace. 
Instead of preserving text CoT as an independent inference-time path, UniVLR renders reasoning traces together with auxiliary images and learns to compress this unified representation into compact visual latent tokens. 
At inference time, the model reasons only through visual latents and directly decodes the final answer, avoiding both external tool calls and verbose text reasoning. 
Experiments on real-world perception and visual reasoning tasks show that UniVLR outperforms prior visual latent reasoning methods while using substantially fewer generated reasoning tokens, suggesting a more unified and efficient paradigm for visual thinking in MLLMs.
 Our code is available at: \url{https://github.com/Warrenustc1958/UniVLR}.
\end{abstract}

\section{Introduction}

Multimodal large language models (MLLMs) are moving beyond \emph{thinking about images} toward \emph{thinking
with images} \cite{think_with_image}. 
They are increasingly deployed in perception-heavy scenarios such as geometric problem solving, chart question answering, high-resolution visual understanding, and embodied planning, where the model must repeatedly attend to, localize, and integrate visual information throughout the reasoning process \cite{V*,HRBench,mathvista,MME}. 
To support this capability, the tool-based visual reasoning paradigm augments the reasoning chain with externally generated auxiliary images, such as crops, annotations, sketches, or intermediate diagrams, and interleaves them with textual chain-of-thought (CoT) \cite{tool_1,tool_2,deepeyes,pixel}. 
While effective, this paradigm is constrained by the rigidity of predefined tools, the latency of external calls, and the instability of generated auxiliary images. 
These limitations make it difficult to support flexible and continuous visual reasoning. 
To overcome this bottleneck, \textbf{visual latent reasoning} has recently emerged as a promising alternative: instead of invoking external tools, the model autoregressively generates latent visual tokens in the visual embedding space, enabling a form of internal visual reasoning that is not tied to explicit tool calls \cite{mirage,LVR,monet,skila,dmlr}.

\begin{figure*}[t]
    \centering
    \includegraphics[width=\linewidth, trim={0 0 0 0}, clip]{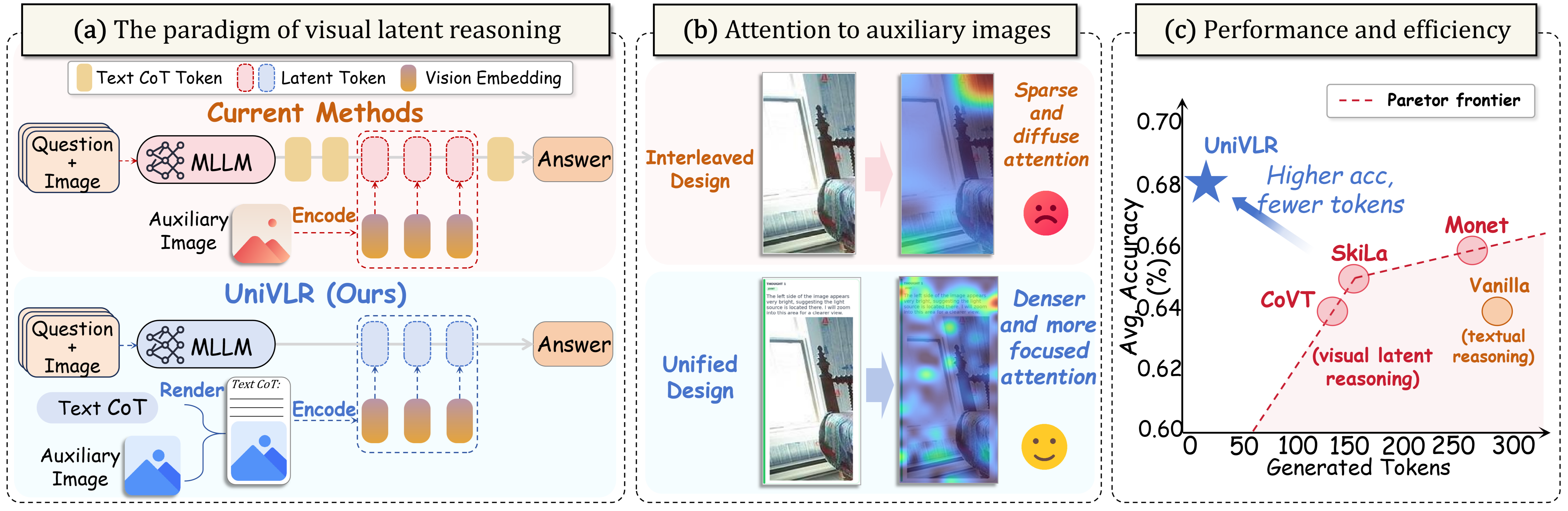}
    \vspace{-6pt}
    \caption{
    \textbf{ Comparison between the current methods and UniVLR.}
    (a) Paradigm illustration comparing the interleaved design of existing visual latent reasoning methods with UniVLR.
    (b) Attention visualization showing that UniVLR induces denser and more focused attention to auxiliary images than the interleaved design.
    (c) Accuracy--efficiency comparison on visual reasoning benchmarks, where UniVLR achieves higher average accuracy with fewer generated tokens. Best viewed in color.
    }
    \label{fig:intro}
    \vspace{-10pt}
\end{figure*}

Despite recent progress, most existing visual latent reasoning methods adopt an interleaved design, where explicit text CoT and latent visual tokens are generated alternately, as shown in Figure~\ref{fig:intro}(a) \cite{mirage,LVR,monet,skila,dmlr,COVT}. 
This design has two practical limitations. 
First, the efficiency gain is limited. 
A key motivation of latent reasoning is to shorten the reasoning trajectory, yet the interleaved paradigm still requires full explicit text CoT segments between visual latent steps \cite{coconut,soft-think,latent-survey}. 
As a result, the overall generation cost is not substantially reduced compared with standard text-based CoT. 
Second, and more importantly, auxiliary images may not fully participate in reasoning. 
Because explicit text tokens and latent visual tokens appear in heterogeneous forms within a long interleaved sequence, the model can easily rely on nearby text tokens while failing to consistently attend to earlier auxiliary images. 
Our analysis further shows that, under interleaved design, subsequent CoT tokens exhibit sparse and diffuse attention to auxiliary images, as illustrated in Figure~\ref{fig:intro}(b). 
This suggests that, in the interleaved setting, the model may still rely heavily on explicit textual reasoning, while auxiliary images are not always fully integrated into subsequent reasoning steps.

\textbf{Does visual latent reasoning really need explicit text CoT as a separate reasoning channel?}
In this paper, we propose a different perspective: explicit text CoT is not necessarily an independent reasoning path that must be preserved; instead, it can be visualized and organized together with auxiliary images within a unified representation space.
This view is motivated by the cognitive observation that reading is visually grounded, and in complex reasoning, humans often organize notes, annotations, and diagrams within a shared perceptual workspace, rather than switching between separate text and image channels \cite{human_1,human_2}.
Likewise, modern MLLMs are equipped with vision encoders that have acquired strong OCR and layout understanding abilities through multimodal pretraining \cite{deepseek-ocr,deepseek-ocr2,qwen2.5-vl,qwen3-vl,onelatent,render-of-thought}, making rendered text a readily interpretable visual signal.
Therefore, intermediate textual reasoning does not have to remain as explicit text tokens; it can be rendered into images, spatially composed with auxiliary images, and processed through the same visual pathway.
Empirically, we observe that such unified design leads to denser and more focused attention over auxiliary images, with key regions being more consistently attended, as shown in Figure~\ref{fig:intro}(b).
This suggests that a unified visual representation not only preserves textual reasoning, but also enables auxiliary images to more actively participate in the reasoning process.

Based on this observation, we introduce \textbf{UniVLR}, a new visual latent reasoning paradigm that represents both textual reasoning and auxiliary images within a single unified visual representation, as shown in Figure~\ref{fig:intro}(a). 
UniVLR is trained in two stages. 
In the first stage, \emph{Visual Latent Grounding}, we use visual CoT auxiliary images as latent alignment targets, enabling the model to autoregressively generate semantically meaningful latent visual tokens in the visual embedding space. 
In the second stage, \emph{Text--Vision Unified Alignment}, we render each textual reasoning step into an image and spatially concatenate it with the corresponding auxiliary image, producing a unified image that jointly carries the reasoning process and the auxiliary image. 
The vision encoder representation of this unified image is then used as the latent alignment target, allowing the model to learn to conduct the entire reasoning process through unified visual latent tokens. 
At inference time, the reasoning trajectory is composed entirely of unified visual latent tokens, and only the final answer is decoded in natural language. 
With this simple yet effective design, UniVLR can organize multi-step reasoning, multiple reference images, and complex annotations on a unified visual canvas, enabling efficient visual latent reasoning without maintaining explicit text CoT as a separate channel.

We conduct extensive experiments on real-world perception and reasoning benchmarks, including V* \cite{V*}, HRBench \cite{HRBench}, and MME-RealWorld \cite{MME}, with multiple base MLLMs such as Qwen2.5-VL \cite{qwen2.5-vl}.
Remarkably, even after removing explicit text CoT during inference, UniVLR outperforms interleaved visual latent reasoning methods that preserve the text channel, including Monet, LVR, and SkiLa. 
On average, UniVLR improves reasoning accuracy by \textbf{5.4\%} while reducing the number of generated tokens by \textbf{15.2$\times$}, as shown in Figure~\ref{fig:intro}(c). 
Moreover, the hidden-state distributions of UniVLR's latent tokens are more coherent, suggesting that textual reasoning and auxiliary images are better aligned within a shared visual latent space. 
The consistent gains across different base MLLMs further demonstrate the robustness and generality of UniVLR, pushing visual latent reasoning toward a more unified form of multimodal thinking.

\section{Method}
\label{sec:method}

\begin{figure*}[t]
    \centering
    \includegraphics[width=\linewidth, trim={0 0 0 0}, clip]{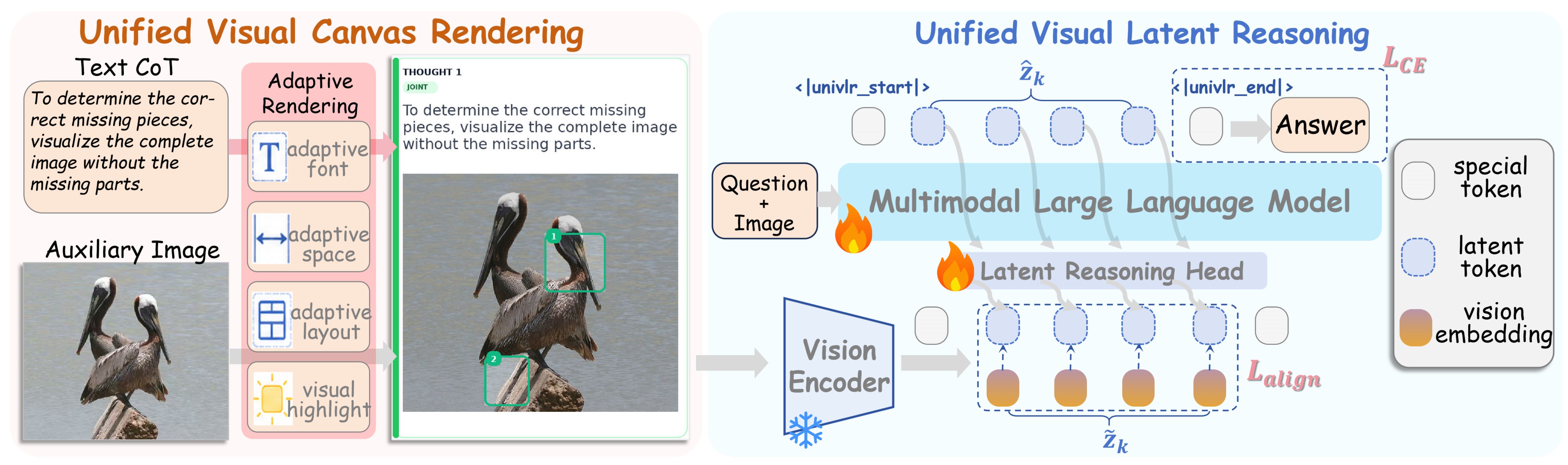}
    \vspace{-6pt}
    \caption{
\textbf{Overview of the proposed UniVLR.}
Left: Unified Visual Canvas Rendering.
Textual reasoning traces are rendered and composed with auxiliary visual evidence into a shared canvas, enabling both reasoning semantics and visual evidence to be encoded by the same vision encoder.
Right: Unified Visual Latent Alignment.
The canvas embeddings are compressed into fixed-length latent targets, and the MLLM learns to autoregressively generate continuous visual latent tokens before decoding the final answer.
}
    \label{fig:method}
    \vspace{-10pt}
\end{figure*}

In this section, we present \textbf{UniVLR}, a unified visual latent reasoning framework that represents textual reasoning and auxiliary images within a single visual latent channel, as illustrated in Figure \ref{fig:method}. 
Section~\ref{subsec:canvas_rendering} introduces unified visual canvas rendering, which converts explicit text CoT and auxiliary visual evidence into a shared visual representation. 
Section~\ref{subsec:latent_training} describes the two-stage latent alignment procedure, including Visual Latent Grounding and Text--Vision Unification. 
Section~\ref{subsec:inference} presents the continuous autoregressive inference process, where the model reasons with compact unified visual latent tokens and decodes only the final answer in natural language.

\subsection{Unified Visual Canvas Rendering}
\label{subsec:canvas_rendering}

In this section, we describe how UniVLR converts heterogeneous reasoning traces into a unified visual representation. Existing visual latent reasoning methods usually preserve explicit text CoT as a separate discrete sequence while using auxiliary images as visual latent supervision. 
In contrast, UniVLR renders textual reasoning and auxiliary images into the same visual workspace, allowing both sources of information to be processed through the vision pathway.

\textbf{Unified canvas construction.}
For each training instance, let $\mathcal{R}=\{r_1,\ldots,r_L\}$ denote the textual reasoning trace and $\mathcal{U}=\{u_1,\ldots,u_M\}$ denote the corresponding auxiliary images, such as crops, annotations, sketches, or intermediate diagrams. 
We define a rendering function $\Phi(\cdot)$ that composes them into a unified visual canvas:
\begin{equation}
    c = \Phi(\mathcal{R}, \mathcal{U}).
\end{equation}
The rendered canvas preserves the semantic content of text CoT while spatially organizing it with auxiliary visual evidence. 
This converts explicit text reasoning from a separate language channel into a visually readable signal.

\textbf{Adaptive rendering.}
To make the canvas compact and readable, $\Phi(\cdot)$ applies lightweight adaptive rendering strategies. 
For the textual part, it adjusts font size, line wrapping, and spacing according to the length of each reasoning step. 
For the visual part, it resizes auxiliary images according to their aspect ratios and optionally highlights salient regions with simple visual cues such as boxes or arrows. 
These operations are not task-specific tools, but a general interface for representing text and visual reasoning on the same canvas. More details of the rendering algorithm are provided in Appendix \ref{app:rendering_algo}.

\textbf{Target visual embeddings.}
We use the vision encoder of the base MLLM to extract a 2D visual feature map from the unified canvas:
\begin{equation}
    \mathbf{F}=E_{\mathrm{vis}}(c)\in \mathbb{R}^{H_f\times W_f\times d},
\end{equation}
where $d$ is the hidden dimension of the language model.
To obtain exactly $K$ latent supervision targets while preserving the spatial structure of the canvas, we adopt aspect-aware 2D pooling.
We choose a pooling grid $(h^\star,w^\star)$ satisfying $h^\star w^\star=K$ by minimizing the aspect-ratio distortion:
\begin{equation}
(h^\star,w^\star)
=
\arg\min_{h w = K}
\left|
\log \frac{w}{h}
-
\log \frac{W_f}{H_f}
\right|.
\end{equation}
The feature map is then adaptively pooled with this grid and flattened into the target latent sequence:
\begin{equation}
    \tilde{\mathbf{Z}}
    =
    \Pi_{h^\star,w^\star}(\mathbf{F})
    =
    [\tilde{\mathbf{z}}_1,\ldots,\tilde{\mathbf{z}}_K]
    \in \mathbb{R}^{K\times d}.
\end{equation}
The resulting embeddings preserve the coarse layout of the rendered reasoning canvas and serve as visual supervision targets for latent alignment.

\subsection{Unified Visual Latent Alignment}
\label{subsec:latent_training}

After constructing visual supervision targets, we train the model to autoregressively generate continuous visual latent tokens.
Instead of reconstructing explicit text CoT, our goal is to align the model's latent reasoning states with visual embeddings that encode intermediate reasoning evidence.
We use a two-stage alignment strategy to make this process stable and semantically unified.

\paragraph{Stage I: Visual Latent Grounding.}
The first stage establishes a basic continuous reasoning interface between the MLLM and its visual embedding space.
We use auxiliary visual reasoning images as latent alignment targets.
Given an auxiliary image, we extract its visual embeddings and compress them into a fixed-length latent sequence
$\tilde{\mathbf{Z}}^{(1)}=\{\tilde{\mathbf{z}}^{(1)}_k\}_{k=1}^{K}$.
The model is then trained to enter latent reasoning mode, autoregressively generate $K$ continuous latent tokens, and return to answer decoding.

\paragraph{Stage II: Text--Vision Unified Alignment.}
The second stage performs the key unification step.
Starting from the Stage-I checkpoint, we replace auxiliary-image targets with unified canvas targets.
Each canvas contains both rendered textual reasoning and auxiliary visual evidence, and its visual embeddings are compressed into
$\tilde{\mathbf{Z}}^{(2)}=\{\tilde{\mathbf{z}}^{(2)}_k\}_{k=1}^{K}$.
Training on these targets teaches the model to represent explicit reasoning traces and visual evidence through the same latent visual channel.

\paragraph{Latent alignment Objective.}
Both stages use the same autoregressive latent training format:
\begin{equation}
    \mathcal{S}
    =
    [\mathcal{X},
    \texttt{<|univlr\_start|>},
    \tilde{\mathbf{z}}_1,\ldots,\tilde{\mathbf{z}}_K,
    \texttt{<|univlr\_end|>},
    \mathcal{A}],
\end{equation}
where $\mathcal{X}$ is the original multimodal input, $\mathcal{A}$ is the final answer, and
$\tilde{\mathbf{z}}_k$ denotes the target latent embedding from either Stage I or Stage II.
At the $k$-th latent step:
\begin{equation}
    \hat{\mathbf{z}}_k = g_{\phi}(\mathbf{h}_{k-1,i}).
\end{equation}
We align predicted and target latent tokens with a normalized regression loss:
\begin{equation}
\mathcal{L}_{\mathrm{align}}
=
\frac{1}{K}
\sum_{k=1}^{K}
\left[
\frac{1}{d}
\left\|
\operatorname{LN}(\hat{\mathbf{z}}_k)
-
\operatorname{LN}(\tilde{\mathbf{z}}_k)
\right\|_2^2
+
1-
\cos\left(
\operatorname{LN}(\hat{\mathbf{z}}_k),
\operatorname{LN}(\tilde{\mathbf{z}}_k)
\right)
\right].
\end{equation}

The final objective combines latent alignment with standard language modeling:
\begin{equation}
    \mathcal{L}
    =
    \mathcal{L}_{\mathrm{CE}}
    +
    \lambda_{\mathrm{align}}\mathcal{L}_{\mathrm{align}}.
\end{equation}
Here, $\mathcal{L}_{\mathrm{CE}}$ is applied only to discrete tokens, including the two special control tokens \texttt{<|univlr\_start|>} and \texttt{<|univlr\_end|>}, as well as the final answer tokens in $\mathcal{A}$.
During training, we apply latent teacher forcing by feeding the target latent embedding as the next latent input, which stabilizes continuous autoregressive learning.

\subsection{Continuous Autoregressive Inference}
\label{subsec:inference}

After training, UniVLR performs inference without rendering text CoT or extracting target visual embeddings. 
The rendering function $\Phi(\cdot)$, the vision encoder used for target extraction, and all teacher latent targets are discarded. 
The model receives only the original multimodal prompt $\mathcal{X}=(\mathcal{Q},\mathcal{I})$.

\textbf{Latent reasoning mode.}
Once the model generates \texttt{<|univlr\_start|>}, it switches from discrete token generation to continuous latent reasoning. 
At latent step $k$, the model predicts a latent token from the previous certain layer $i$ hidden state $\mathbf{h}_{k-1,i}$:
   $ \hat{\mathbf{z}}_k = g_{\phi}(\mathbf{h}_{k-1,i}).$
The predicted latent token is directly fed back as the input embedding for the next step:
\begin{equation}
    \mathbf{e}_{k} = \hat{\mathbf{z}}_k.
\end{equation}
This recurrence is repeated for a predefined latent budget of $K$ steps.

\textbf{Answer decoding.}
After $K$ latent steps, the model emits \texttt{<|univlr\_end|>} and switches back to the discrete vocabulary space to decode the final answer $\mathcal{A}$. 
Therefore, inference consists of a compact continuous reasoning phase followed by a short natural-language answer phase. 
This allows UniVLR to retain the information of text CoT and auxiliary images without generating explicit intermediate reasoning tokens at test time.
\section{Experiments}
\label{sec:experiments}

In this section, we conduct extensive experiments to address the following research questions:

\textbf{RQ1: }
How does UniVLR perform compared with baselines, and can it achieve competitive or superior performance with substantially fewer reasoning tokens?

\textbf{RQ2: }
How are the hidden-state representations of UniVLR's latent reasoning tokens distributed compared with explicit text tokens and image tokens? Do the learned latent tokens align more closely with the visual-token representation space?

\textbf{RQ3: }
How does the number of latent reasoning tokens affect the performance and efficiency of UniVLR, and what latent token budget provides the best balance?

\textbf{RQ4: }
How do key components of UniVLR, including visual-text rendering, unified canvas construction, contribute to the final performance?

\subsection{Experimental Setup}
In this subsection, we summarize the evaluation benchmarks, dataset, base models and baseline
methods used in our experiments. Further details and additional experiments are provided in Appendix \ref{app:implementation_details} and Appendix \ref{app:additional_experiment_results}.

\textbf{Benchmarks \& Metrics.}
We evaluate UniVLR on a diverse suite of perception-centric and visual reasoning benchmarks.
Specifically, we use \textbf{V*} \cite{V*} to evaluate guided visual search and spatial reasoning, \textbf{HRBench4K} \cite{HRBench} and \textbf{HRBench8K} \cite{HRBench} to assess high-resolution fine-grained perception, and \textbf{MME-RealWorld-Lite} \cite{MME} to measure real-world multimodal comprehension.
For each benchmark, we report the official overall score as well as its corresponding sub-category scores when available.
In addition, we use the average of the overall scores across benchmarks as a compact indicator of general performance.

\textbf{Base models \& Baselines.}
We compare UniVLR with representative methods from three reasoning paradigms.
First, \textit{textual reasoning} baselines include GPT-4o, Qwen2.5-VL-7B, and a vanilla SFT variant of Qwen2.5-VL-7B, which rely primarily on explicit textual reasoning or direct answer generation.
Second, \textit{tool-based visual reasoning} methods, including PixelReasoner \cite{pixel} and DeepEyes \cite{deepeyes}, improve perception by invoking external visual operations such as cropping, zooming, or multi-turn image inspection.
Third, \textit{visual latent reasoning} methods include LVR \cite{LVR}, Monet \cite{monet}, SkiLa \cite{skila},CoVT\cite{COVT} which introduce latent visual reasoning tokens while still preserving explicit textual reasoning channels. 
Unlike these methods, UniVLR removes the explicit reasoning text channel at inference time and performs reasoning through a compact sequence of unified visual latent tokens.

\begin{table*}[!htbp]
    \centering
    \caption{
    \textbf{Comparison with representative baselines on perception-centric and visual reasoning benchmarks.}
     The best results are highlighted in \textbf{bold}. Results marked with `*' are taken from prior work. UniVLR achieves strong overall performance across all evaluated benchmarks.}
    \label{tab:main_results}

    \begingroup
    \setlength{\tabcolsep}{3pt}
    \renewcommand{\arraystretch}{1.08}
    \normalsize

    \begin{adjustbox}{max width=\textwidth}
    \begin{tabular}{lcccccccccccc}
        \toprule
        \multirow{2}{*}{\textbf{Model}} 
        & \multicolumn{3}{c}{\textbf{V*}} 
        & \multicolumn{3}{c}{\textbf{HRBench4K}} 
        & \multicolumn{3}{c}{\textbf{HRBench8K}} 
        & \multicolumn{3}{c}{\textbf{MME-RealWorld-Lite}} \\
        \cmidrule(lr){2-4}
        \cmidrule(lr){5-7}
        \cmidrule(lr){8-10}
        \cmidrule(lr){11-13}
        & Overall & Attr. & Spa. 
        & Overall & FSP & FCP 
        & Overall & FSP & FCP 
        & Overall & Rea. & Perc. \\
        
        \midrule
        \rowcolor[HTML]{FEE090}
        \multicolumn{13}{c}{\textbf{Textual Reasoning}}\\
        GPT-4o~\cite{gpt_4o}
        & 67.5* & 72.2* & 60.5* 
        & 59.0* & 70.0* & 48.0* 
        & 55.5* & 62.0* & 49.0* 
        & 52.0* & 48.3* & 54.4* \\

        Qwen2.5-VL-7B~\cite{qwen2.5-vl}
        & 77.4 & 78.3 & 76.3 
        & 69.0 & 85.8 & 52.2 
        & 66.0 & 80.3 & 51.8 
        & 46.2 & 43.1 & 48.2 \\
        
        \quad + vanilla SFT 
        & 75.4 & 80.0 & 68.5 
        & 69.1 & 81.3 & 57.0 
        & 63.6 & 73.3 & 54.0 
        & 45.5 & 39.9 & 49.1 \\
        
        \midrule
        \rowcolor[HTML]{E0F3F8}
        \multicolumn{13}{c}{\textbf{Tool-based Visual Reasoning}}\\
        \rowcolor[HTML]{F0F0F0}
        PixelReasoner~\cite{pixel}
        & {\color{black!70}80.6*} & {\color{black!70}83.5*} & {\color{black!70}76.3*} 
        & {\color{black!70}72.9*} & {\color{black!70}86.0*} & {\color{black!70}60.3*} 
        & {\color{black!70}66.9*} & {\color{black!70}80.0*} & {\color{black!70}54.3*} 
        & {\color{black!70}49.7*} & {\color{black!70}44.5*} & {\color{black!70}53.1*} \\
        
        \rowcolor[HTML]{F0F0F0}
        DeepEyes ~\cite{deepeyes}
        & {\color{black!70}83.3*} & {\color{black!70}84.4*} & {\color{black!70}81.6*} 
        & {\color{black!70}71.3*} & {\color{black!70}83.8*} & {\color{black!70}58.8*} 
        & {\color{black!70}65.1*} & {\color{black!70}77.0*} & {\color{black!70}53.3*} 
        & {\color{black!70}54.3*} & {\color{black!70}50.5*} & {\color{black!70}56.6*} \\
    
        \midrule
        \rowcolor[HTML]{91BFDE}
        \multicolumn{13}{c}{\textbf{Visual Latent Reasoning}}\\
        LVR ~\cite{LVR}
        & 80.6 & 81.7 & 79.8 
        & 69.9 & 84.7 & 55.7 
        & 66.9 & 77.7 & 56.2 
        & 39.1 & 37.5 & 40.1 \\
        Monet ~\cite{monet}
        & 79.1 & 81.7 & 75.0
        & 71.9 & \textbf{89.3} & 54.5 
        & 63.5 & 76.5 & 50.5 
        & 46.9 & 40.3 & 51.2 \\
        SkiLa~\cite{skila}
        & 80.1 & 79.1 & 81.6 
        & 70.3 & 84.7 & 55.7 
        & 62.9 & 77.5 & 48.3 
        & 45.6 & 36.3 & 51.6 \\
        CoVT~\cite{COVT}
        & 78.0 & 79.1 & 76.3 
        & 71.9 & 84.2 & 59.5 
        & \textbf{69.7} & \textbf{85.4} & 54.0 
        & 48.2 & 42.9 & 51.6 \\

        \midrule
        \textbf{UniVLR}
        & \textbf{82.7} & \textbf{83.5} & \textbf{81.6} 
        & \textbf{73.3} & 86.0 & \textbf{60.5}
        & 68.8 & 78.8 & \textbf{58.8} 
        & \textbf{50.7} & \textbf{44.7} & \textbf{54.5} \\
        
        \bottomrule
    \end{tabular}%
    \end{adjustbox}

    \endgroup

    \vspace{2pt}
    \vspace{-5pt}
\end{table*}
\textbf{Implementation Details.}
We instantiate UniVLR on top of Qwen2.5-VL-7B-Instruct \cite{qwen2.5-vl}.
To preserve the pretrained visual prior, we freeze the vision encoder and patch-merger, and fine-tune the LLM backbone together with a lightweight MLP-based latent reasoning head.
\textbf{Unless otherwise specified, the teacher forcing latent budget is fixed to $K_{train}=24$. At inference time, we use a shorter latent reasoning budget by default, setting $K_{infer}=12$.}
We report both UniVLR-Stage1 and the final UniVLR model, where the former serves as an intermediate variant and the latter denotes our final model used for main comparison.
Training details, including optimization hyperparameters and data construction, are provided in Appendix~\ref{app:implementation_details}.

\subsection{Main Results and Efficiency Analysis (RQ1)}
\label{exp:main_results}

To answer RQ1, we compare UniVLR (UniVLR-Stage2) with textual reasoning, tool-based visual reasoning, and visual latent reasoning baselines on four main benchmarks. UniVLR's latent reasoning budget is fixed to $K_{\text{infer}}=12$. This shorter inference budget is chosen according to the token-scaling analysis in Section\ref{exp:latent_token_number}.
Table~\ref{tab:main_results} reports the performance comparison, and Figure~\ref{fig:rq2}(a) further compares the reasoning-token budget.
Based on the results, we find that:

\begin{itemize}[leftmargin=*]
    \item \textbf{Obs 1: UniVLR improves over textual reasoning baselines.}
    Compared with Qwen2.5-VL-7B, UniVLR improves the overall scores from 77.4 to 82.7 on V*, 69.0 to 73.3 on HRBench4K, 66.0 to 68.8 on HRBench8K, and 46.2 to 50.7 on MME-RealWorld-Lite.
    In contrast, vanilla SFT does not yield consistent gains, indicating that the improvement comes from unified visual latent reasoning rather than simple fine-tuning.

    \item \textbf{Obs 2: UniVLR achieves the strongest overall performance among visual latent reasoning methods.}
    UniVLR obtains the best overall scores on V*, HRBench4K, and MME-RealWorld-Lite.
    The average overall score of UniVLR reaches 68.9, outperforming LVR, Monet, and SkiLa.
    This suggests that unifying text traces and auxiliary images into one visual latent channel is more effective than interleaved text--latent reasoning.

    \item \textbf{Obs 3: UniVLR substantially reduces generated reasoning tokens by removing explicit intermediate text generation.} 
    As depicted in Figure~\ref{fig:rq2}(a), interleaved methods (\eg Monet, SkiLa, and CoVT) generate 190 to 270 reasoning tokens per instance, predominantly explicit text. Even LVR, which only enhances visual grounding, requires explicit text CoT. In stark contrast, UniVLR achieves competitive or stronger performance using only 12 latent tokens and no generated intermediate text tokens. This reduction in generated reasoning length suggests that UniVLR can encode useful intermediate reasoning information in a compact continuous format.

\end{itemize}

\subsection{Unified Latent Representation Analysis (RQ2)}
\label{exp:efficiency_and_latent_distribution}

\begin{figure*}[htbp]
    \centering
    \includegraphics[width=\linewidth, trim={0 0 0 0}, clip]{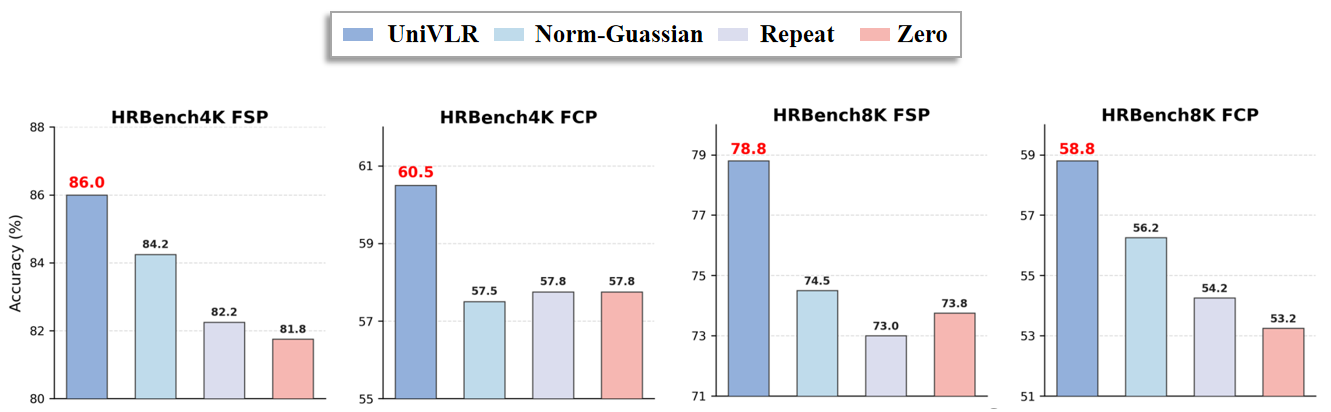}
    \vspace{-6pt}
    \caption{
    \textbf{Causal perturbation analysis of latent reasoning tokens.}
    We perturb latent hidden states during inference by zeroing them, injecting Gaussian noise, or repeating the first latent state.
    The resulting accuracy drops indicate that latent tokens carry task-relevant reasoning information.
    }
    \label{fig:perturbation}
    \vspace{-5pt}
\end{figure*}

To answer RQ2, we analyze whether the generated latent tokens behave as meaningful visual reasoning states.
We examine their token efficiency, last-hidden-state distribution, and sensitivity to inference-time perturbations.
The results are shown in Figures~\ref{fig:rq2} and \ref{fig:perturbation}, from which we can find that: 

\begin{figure*}[htbp]
    \centering
    \includegraphics[width=\linewidth, trim={0 0 0 0}, clip]{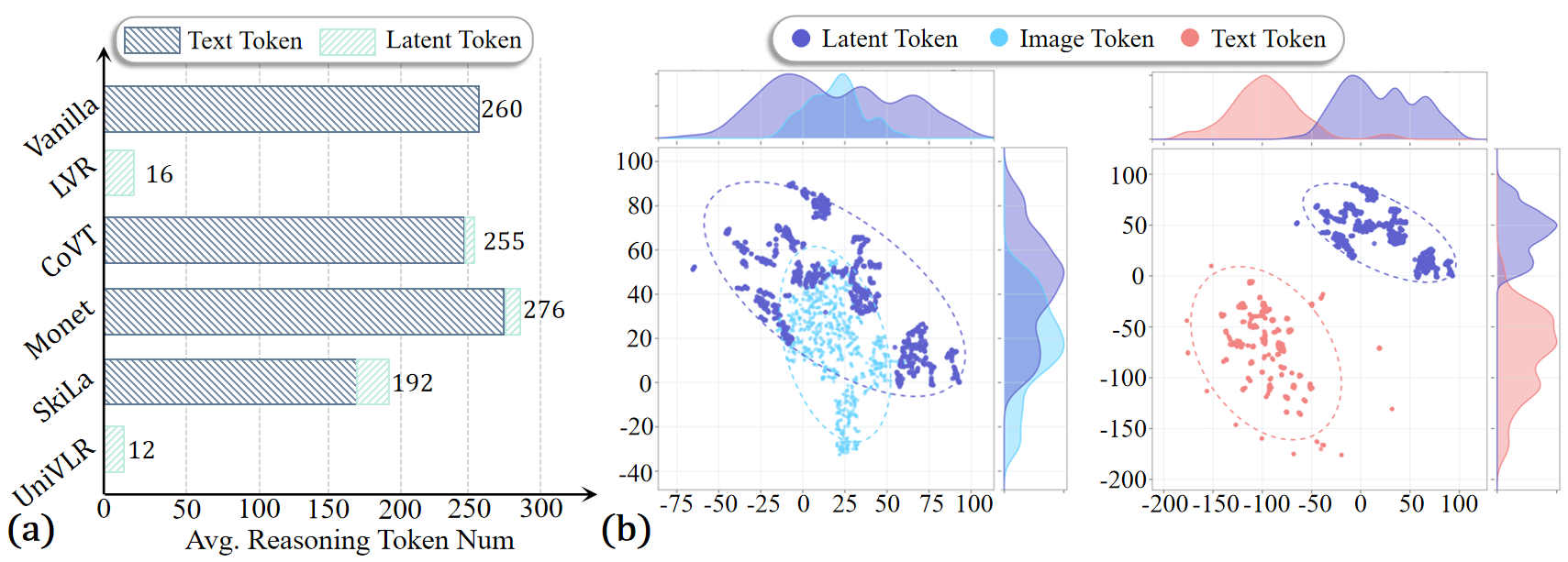}
    \vspace{-6pt}
    \caption{
    \textbf{Reasoning-token efficiency and latent representation distribution.}
    \textbf{(a)} UniVLR performs reasoning with a compact latent-token budget and no generated text CoT.
    \textbf{(b)} Last-hidden-state visualization shows that UniVLR latent tokens are closer to image-token representations than to text-token representations.
    }
    \label{fig:rq2}
    \vspace{-10pt}
\end{figure*}

\begin{itemize}[leftmargin=*]
    \item \textbf{Obs 4: UniVLR latent tokens align more closely with visual representations than textual ones.}
    In Figure~\ref{fig:rq2}(b), latent tokens overlap substantially with image-token hidden states, while remaining clearly separated from text-token states.
    This suggests that UniVLR’s latent states are closer to visual-token representations, providing evidence that the model uses a visually grounded latent channel rather than only an implicit textual CoT channel.

    \item \textbf{Obs 5: Latent tokens have a direct impact on final answer prediction.}
    In Figure~\ref{fig:perturbation}, all perturbations consistently reduce accuracy on HRBench.
    For example, zeroing latent states drops HRBench8K-FCP from 58.8 to 53.2, and repeating the first latent state drops HRBench8K-FSP from 78.8 to 73.0.
    This indicates that the generated latent sequence contains task-relevant information and is not merely an unused placeholder.
\end{itemize}

\subsection{Effect of Latent Token Number (RQ3)}
\label{exp:latent_token_number}

\label{exp:latent_distribution}
\begin{figure*}[htbp]
    \centering
    \includegraphics[width=\linewidth, trim={0 0 0 0}, clip]{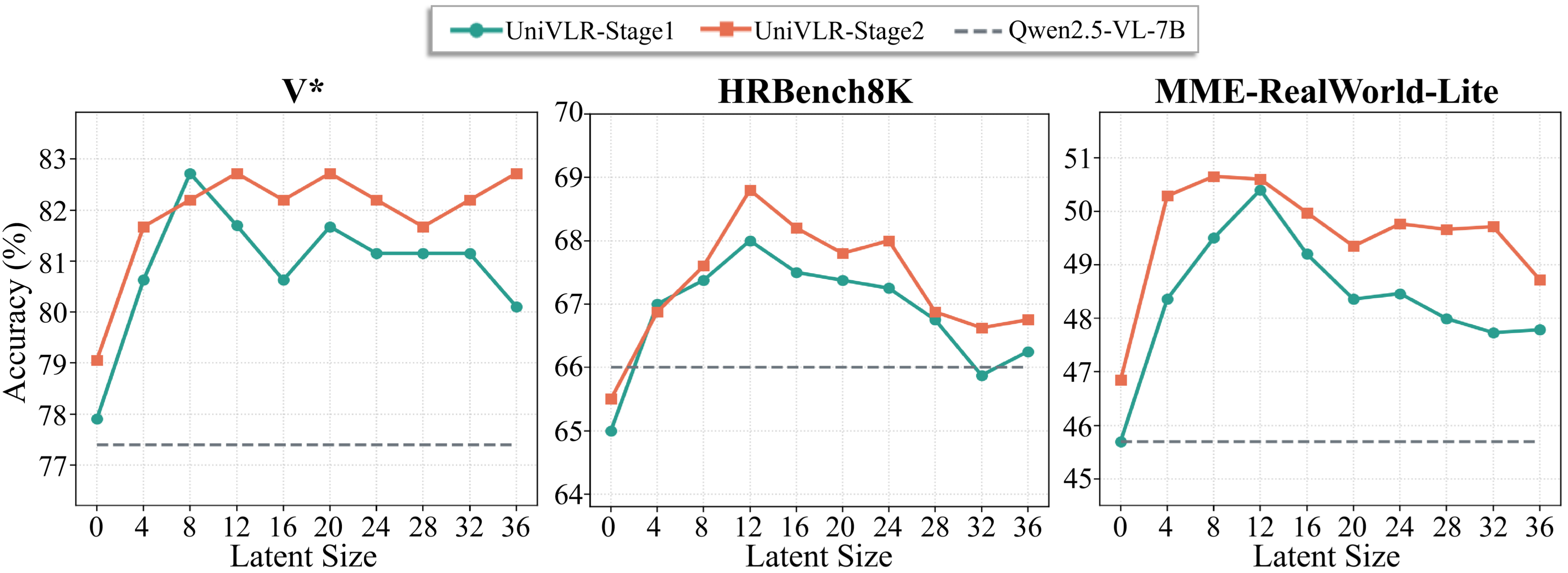}
    \vspace{-6pt}
    \caption{
\textbf{Effect of the number of unified visual latent tokens during inference on test accuracy.} Both UniVLR-Stage1 and UniVLR are trained with a fixed teacher-forcing latent size of $K=24$. The dashed line marks the zero-shot accuracy of  Qwen2.5-VL-7B.
}
    \label{fig:latent_scaling}
    \vspace{-10pt}
\end{figure*}

\begin{table*}[htbp]
    \centering
    \caption{
    \textbf{Comprehensive Ablation Study decoupled by training stages.} 
    \textbf{(Left)} REPA ablations evaluated on Stage I models. 
    \textbf{(Right)} RFC ablations evaluated on Stage II models. 
    Both stages are evaluated consistently across V*, HRBench8K, and fine-grained subsets of MME-RealWorld-Lite. Our default UniVLR settings are highlighted in \colorbox{gray!20}{gray}.
    }
    \label{tab:ablation}
    \vspace{2pt}
    \renewcommand{\arraystretch}{1.12}
    \setlength{\tabcolsep}{2.8pt}
    \large
    \resizebox{\textwidth}{!}{
        
        \begin{tabular}[t]{l cc ccc} 
            \toprule
            \multirow{2}{*}{\textbf{Stage I: REPA Ablations}} & \textbf{V*} & \textbf{HRBench8K} & \multicolumn{3}{c}{\textbf{MME-RealWorld-Lite}} \\
            \cmidrule(lr){2-2} \cmidrule(lr){3-3} \cmidrule(lr){4-6}
            & Overall & Overall & \textbf{Overall} & Rea. & Perc. \\
            \midrule
            \rowcolor{gray!20} \textbf{UniVLR-Stage1 (Default)} & \textbf{81.7} & \textbf{68.0} & \textbf{50.4} & 42.4 & \textbf{55.6} \\
            \midrule
            \multicolumn{6}{c}{\textit{Extraction Strategy (Default: Avg 2D Pool)}} \\
            \midrule
            Avg 1D Pool & 79.1 & 66.3 & 46.5 & 38.0 & 51.9 \\
            MLERP 2D Pool & 71.2 & 59.1 & 48.5 & 41.0 & 53.2 \\
            \midrule
            \multicolumn{6}{c}{\textit{Latent Reasoning Head (Default: MLP)}} \\
            \midrule
            w/o Head & 81.2 & 66.5 & 48.7 & \textbf{43.2} & 52.2 \\
            GLU Head & 81.2 & 64.4 & 48.4 & 41.9 & 52.5 \\
            \midrule
            \multicolumn{6}{c}{\textit{Hidden States Aligned Layer (Default: Middle)}} \\
            \midrule
            Front Layer & 81.5 & 66.3 & 49.2 & 40.7 & 54.7 \\
            Last Layer & 81.2 & 67.3 & 48.5 & 41.2 & 53.1 \\
            \bottomrule
        \end{tabular}

        \hspace{0.006\textwidth} 

        \begin{tabular}[t]{l cc ccc} 
            \toprule
            \multirow{2}{*}{\textbf{Stage II: RFC Ablations}} & \textbf{V*} & \textbf{HRBench8K} & \multicolumn{3}{c}{\textbf{MME-RealWorld-Lite}} \\
            \cmidrule(lr){2-2} \cmidrule(lr){3-3} \cmidrule(lr){4-6}
            & Overall & Overall & \textbf{Overall} & Rea. & Perc. \\
            \midrule
            \rowcolor{gray!20} \textbf{UniVLR-Stage2 (Default)} & \textbf{82.7} & \textbf{68.8} & \textbf{50.7} & 44.7 & 54.5 \\
            \midrule
            \multicolumn{6}{c}{\textit{Training Curriculum (Default: Two Stages)}} \\
            \midrule
            Stage I Only (Warm-up) & 81.7 & 68.0 & 50.4 & 42.4 & \textbf{55.6} \\
            Single-Stage Mixed & 77.5 & 64.8 & 47.1 & 42.1 & 50.2 \\
            \midrule
            \multicolumn{6}{c}{\textit{Render Strategy (Default: Vertical Layout\ref{alg:vertical})}} \\
            \midrule
            Left-Right Layout\ref{alg:compact_lr} & 82.1 & 67.8 & 50.1 & 43.5 & 54.5 \\
            Fixed Adaptive Wrap\ref{alg:fixed_wrap} & 81.6 & 66.5 & 48.8 & 44.9 & 51.2 \\
            \midrule
            \multicolumn{6}{c}{\textit{Data Ablation (Default: Filtered Zebra + VisCoT)}} \\
            \midrule
            Unfiltered Zebra-CoT & 79.1 & 65.5 & 49.9 & \textbf{47.1} & 51.8 \\
            Pure Visual-CoT & 81.5 & 66.3 & 49.8 & 43.9 & 53.6 \\
            \bottomrule
        \end{tabular}

    } 
\end{table*}
To answer RQ3, we vary the inference-time latent token number $K$ from $0$ to $36$ and evaluate both UniVLR-Stage1 and UniVLR (UniVLR-Stage2). The results are summarized in  Figure~\ref{fig:latent_scaling}, from which we find that:

\begin{itemize}[leftmargin=*]
    \item \textbf{Obs 6: Latent reasoning improves direct answering, but more tokens are not always better.}
    Across benchmarks, $K=0$ gives the weakest performance, while introducing a small number of latent tokens brings clear gains.
    Performance typically peaks at a moderate budget and then fluctuates or drops, suggesting that excessive latent steps may introduce representation drift.

    \item \textbf{Obs 7: Stage-II training makes latent scaling more stable.}
    Compared with UniVLR-Stage1, the final UniVLR generally achieves higher performance across nonzero $K$ values.
    It also shows a wider stable plateau, especially on MME-RealWorld-Lite, indicating that unified text--vision alignment improves robustness to the inference-time latent budget.
\end{itemize}

\subsection{Ablation Study (RQ4)}
\label{exp:ablation}

To answer RQ4, we rigorously disentangle the contributions of our proposed designs. As shown in Table~\ref{tab:ablation}, we decouple the ablation studies into two orthogonal dimensions: \textbf{Representation Alignment (REPA)} evaluated on Stage I, and \textbf{Reasoning Formulation and Curriculum (RFC)} evaluated on Stage II. We summarize our findings below:

\begin{itemize}[leftmargin=*]
    \item \textbf{Obs 8: For RFC, an incremental curriculum and vertical spatial rendering are critical for mastering complex logic.} 
    Directly training on a single-stage mixture of warm-up and hard examples leads to a clear performance drop, with V* decreasing to 77.5. Our two-stage curriculum effectively mitigates representation collapse by anchoring the latent interface first. Furthermore, our \textit{Vertical Layout} strategy outperforms both \textit{Left-Right} and \textit{Fixed Wrap} layouts. Detailed rendering algorithms are provided in Appendix~\ref{app:rendering_algo}. Finally, rigorous data filtering is essential; using unfiltered hard examples introduces noisy or shortcut-prone supervision, which can encourage the latent tokens to encode spurious textual patterns

    \item \textbf{Obs 9: For REPA, preserving 2D spatial topology and utilizing a lightweight projection head are fundamental.} 
    Replacing aspect-aware 2D pooling with Avg 1D Pool or MLERP\cite{kim2024token} weakens the preservation of spatial layout, dropping reasoning performance on MME-RealWorld-Lite e.g., \textit{Avg 1D} drops to 38.0. Architecturally, a simple MLP head optimally balances visual feature alignment and autoregressive stability, outperforming both head-less and over-parameterized \textit{GLU} designs. Moreover, aligning latent tokens to middle-layer hidden states yields better performance than aligning them to final-layer hidden states, likely because middle layers preserve richer visual and spatial information. This finding aligns with previous research (\cite{skean2025layer}\cite{yu2024repa}\cite{lee2025tamp}\cite{kang2025yourvlm})suggesting that the middle layers of MLLMs primarily encode visual information.
\end{itemize}
\section{Related Work}
\label{sec:related_work}

\paragraph{Thinking with Images.}
Recent studies have explored \emph{thinking with images}, where visual information is no longer treated as a passive input but as an active reasoning workspace. 
Representative methods equip MLLMs with visual operations such as cropping, zooming, grounding, or frame selection, and train models to invoke these operations during multi-step reasoning \cite{deepeyes,pixel,tool_1,tool_2,think_with_image}. 
This paradigm is especially effective for fine-grained perception and high-resolution reasoning, since models can actively revisit local visual evidence instead of relying only on the initial image encoding. 
However, such methods usually require explicit tool invocation, multi-turn interaction, or additional visual processing at inference time, which increases latency and makes reasoning dependent on predefined operation sets. 
In contrast, UniVLR does not use external visual tools during inference. 
We instead convert intermediate reasoning evidence into a compact sequence of unified visual latent tokens, allowing the model to internalize visual deliberation within a single latent reasoning channel.

\paragraph{Visual Latent Reasoning.}
Another line of work aims to move intermediate reasoning from discrete text tokens into continuous latent spaces. 
Latent visual reasoning methods train MLLMs to generate visual embeddings, latent sketches, or continuous visual thoughts as intermediate reasoning states, often by reconstructing image features or sketch-like supervision signals \cite{LVR,monet,skila,COVT,dmlr}. 
Recent rendered-reasoning approaches further show that textual CoT can be rendered into images and compressed into visual or latent representations, reducing the cost of verbose textual reasoning while preserving inspectable supervision signals \cite{render-of-thought,onelatent}. 
Despite these advances, existing visual latent reasoning methods usually preserve an explicit textual reasoning channel, alternate between text and latent visual tokens, or focus on compressing textual CoT alone. 
UniVLR differs by unifying rendered textual reasoning and auxiliary visual evidence into a single visual canvas, then aligning autoregressive latent tokens to this unified visual representation. 
Thus, explicit text CoT is not maintained as a separate reasoning path; both reasoning semantics and visual evidence are supervised through a shared compact visual latent space.
\section{Limitations}
\label{sec:limitations}

While \textbf{UniVLR} shows that explicit text CoT can be absorbed into a unified visual latent channel, several limitations remain. 
First, its effectiveness depends on the vision encoder's OCR and layout understanding; models with weaker visual priors may benefit less from the same rendering strategy. 
Second, visual latent tokens are more efficient but less directly inspectable than natural-language rationales. 
Although our representation and perturbation analyses show that they carry meaningful reasoning information, they do not fully provide the transparency of explicit CoT. 
Third, we currently use a fixed latent-token budget, while different tasks may require different amounts of latent computation. 
Finally, UniVLR is not meant to replace tool-based reasoning for tasks requiring exact measurement, exhaustive high-resolution search, or executable visual manipulation, where external tools can remain complementary.

\section{Conclusion}
\label{sec:conclusion}

We introduced \textbf{UniVLR}, a unified visual latent reasoning framework that removes explicit text CoT as a separate inference-time channel. 
By rendering textual reasoning and auxiliary visual evidence into a shared visual workspace, UniVLR learns compact visual latent tokens that carry both reasoning semantics and perceptual information. 
At inference time, it reasons through visual latents and decodes only the final answer, avoiding verbose text reasoning and external tool calls. 
Experiments show that UniVLR improves over prior visual latent reasoning methods with substantially fewer generated tokens, while analyses of hidden states, latent-token scaling, and ablations support the effectiveness of unified visual latent reasoning.
\newpage

\bibliographystyle{unsrt}
\bibliography{references}
\newpage
\appendix
\clearpage
\appendix 


\renewcommand{\thepage}{\arabic{page}}
\renewcommand{\thefigure}{\arabic{figure}}
\renewcommand{\thetable}{\arabic{table}}
\renewcommand{\theequation}{\arabic{equation}}


\section{Broader Impacts}
\label{app:broader_impact}

UniVLR explores a more efficient way for MLLMs to conduct visual reasoning by replacing verbose explicit reasoning traces with compact visual latent tokens. 
This may reduce inference cost and latency for perception-heavy applications such as high-resolution document understanding, chart analysis, visual search, and embodied decision making. 
At the same time, moving reasoning from explicit text into latent visual states may reduce the direct interpretability of intermediate reasoning, which could make debugging, auditing, or safety monitoring more challenging. 
Although our analyses show that the learned latent tokens carry meaningful reasoning information, they do not provide the same human-readable transparency as textual CoT. 
Therefore, future deployments of visual latent reasoning systems should be accompanied by complementary interpretability, monitoring, and failure-diagnosis tools, especially in high-stakes domains where visual reasoning errors may lead to harmful decisions. 
We view UniVLR as a step toward more efficient multimodal reasoning, while emphasizing that efficiency gains should not come at the cost of accountability and safety.
\section{Implementation Details}
\label{app:implementation_details}
\subsection{Benchmark and Model Details}
\label{app:benchmark_model_details}
\paragraph{V*~\cite{V*}}is designed to evaluate MLLMs in their ability to process high-resolution images and focus on visual details.We select 191 samples to complete the evaluation.
\paragraph{HRBench~\cite{HRBench}} designs high-resolution multimodal benchmarks, which consists two sub-tasks: Fine-grained Single-instance Perception (FSP) and Fine-grained Cross-instance Perception (FCP). The FSP task includes 100 samples, which includes tasks such as attribute recognition, OCR, visual prompting. The FCP task also comprises 100 samples which encompasses map analysis, chart analysis and spatial relationship assessment.HR-Bench is available in two versions: HR-Bench 8K and HR-Bench 4K. The HR-Bench 8K includes images with an average resolution of 8K. Additionally, we manually annotate the coordinates of objects relevant to the questions within the 8K image and crop these image to 4K resolution.

\paragraph{MME-RealWorld-Lite~\cite{MME}}contains 13K high-quality images, annotated by 32 volunteers, resulting in 29K question-answer pairs that cover 43 subtasks across 5 real-world scenarios. We choose 1919 samples to evaluate all latent visual reasoning baselines.
\paragraph{Qwen2.5-VL-7B~\cite{qwen2.5-vl}}is a frontier MLLM with exceptional multimodal understanding capabilities and a long context window.
By default, we adopt the natural resolution mechanism to avoid distorting the aspect ratio.
We set a maximum pixel of 5120$\times$28$\times$28 and a minimum pixel of 128$\times$28$\times$28.

\subsection{Baseline Details}
\label{app:baseline_details}
\paragraph{LVR~\cite{LVR}} is a novel multimodal reasoning paradigm that overcomes the limitations of traditional text-only reasoning by enabling autoregressive reasoning directly within the visual embedding latent space. The method first projects images into a joint semantic space shared with the language model, and then trains the LLM to dynamically generate hidden states that reconstruct key visual tokens critical for answering queries. Its training pipeline consists of two stages: Supervised Fine-Tuning  using a MSE loss for visual feature reconstruction, and Reinforcement Learning employing an adapted GRPO algorithm tailored for self-evolution in latent reasoning.
\paragraph{Skila~\cite{skila}}is a unified multimodal reasoning paradigm that expands the autoregressive capabilities of MLLMs to natively generate continuous visual embeddings, termed latent sketch tokens, as visual thoughts. During inference, the method dynamically alternates between a textual thinking mode for generating discrete text and a visual sketching mode for generating continuous features. The training pipeline employs a latent visual semantics reconstruction mechanism, leveraging an auxiliary sketch encoder to extract features from intermediate sketch images as MSE reconstruction targets, ensuring that the generated latent sketch tokens are strictly semantically grounded.
\paragraph{Monet~\cite{monet}}is a training framework that empowers MLLMs to perform abstract reasoning in the latent visual space by generating continuous embeddings as intermediate "visual thoughts," thereby eliminating the reliance on predefined external visual tools. The pipeline first utilizes a three-stage distillation-based Supervised Fine-Tuning process to mitigate the high computational cost of latent-visual alignment. Subsequently, to overcome the inability of standard GRPO to optimize continuous latent features, it introduces VLPO (Visual-latent Policy Optimization), a novel reinforcement learning method that explicitly incorporates latent embeddings into policy gradient updates by estimating their approximate output probabilities.
\paragraph{CoVT~\cite{COVT}}is a framework that guides VLMs to think using continuous visual tokens, aiming to enrich the model with fine-grained, dense visual perception capabilities such as depth, segmentation, and edges. The pipeline trains the VLM to autoregressively predict these continuous visual tokens, aligning them with underlying perceptual signals using lightweight visual expert task decoders e.g. SAM, DepthAnything, PIDINet, DINO guided by reconstruction losses. During inference, the model directly constructs a chain of visual thoughts in the continuous token space, while supporting the optional decoding of these tokens back into dense prediction maps for human interpretability.
\subsection{SFT Training}
\label{app:sft_training}
We build UniVLR on top of the \texttt{Qwen2.5-VL-7B-Instruct} backbone. Throughout all training stages, the pre-trained vision tower and patch-merger are strictly frozen to preserve the original visual representation capabilities. We conduct full-parameter fine-tuning on the LLM backbone and optimize a newly initialized Latent Visual Reasoning projection head. 

\paragraph{UniVLR Projection Head and Alignment Layer.} 
Unlike standard multimodal models that project visual features into the language space at the input layer, we extract the LLM's decoder hidden states to align with the visual target space. We empirically find that aligning intermediate layers e.g., the 14th layer out of 28 layers yields more robust spatial and reasoning features than the final layer. The extracted hidden state $h_k$ is then passed through our UniVLR projection head, which is instantiated as a lightweight Multi-Layer Perceptron (MLP) consisting of: $\text{LayerNorm} \rightarrow \text{Linear} \rightarrow \text{GELU} \rightarrow \text{Linear}$.

\paragraph{Training Sequence and Special Tokens.}
We introduce four special control tokens to regulate the latent reasoning process: \texttt{<|univlr\_start|>}, \texttt{<|univlr|>}, \texttt{<|univlr\_end|>}, and \texttt{<|univlr\_latent\_end|>}. To leverage pretrained priors, we initialize their embeddings using the existing visual boundary tokens e.g. \texttt{<|im\_start|>} for \texttt{<|univlr\_start|>}, \texttt{<|image\_token|>} for \texttt{<|univlr|>}, and \texttt{<|im\_end|>} for the end tokens. 
A complete reasoning trajectory on the assistant side is formatted as:
\begin{equation*}
    \texttt{<|univlr\_start|>} \underbrace{\texttt{<|univlr|>} \dots \texttt{<|univlr|>}}_{K=24} \texttt{<|univlr\_end|>} \texttt{<|univlr\_latent\_end|>} \ \mathcal{A}
\end{equation*}

\paragraph{Masking Strategy for Language Modeling.}
During supervised fine-tuning, the standard Cross-Entropy  loss $\mathcal{L}_{\mathrm{CE}}$ is applied exclusively to the textual answer $\mathcal{A}$ and the structural boundary tokens (\texttt{<|univlr\_start|>} and \texttt{<|univlr\_end|>}), teaching the model when to enter and exit the latent mode. We strictly mask out the prompt tokens, padding tokens, the intermediate \texttt{<|univlr|>} placeholders, and the \texttt{<|univlr\_latent\_end|>} token from the CE loss by setting their labels to \texttt{IGNORE\_INDEX}. 

\paragraph{Latent Alignment Objective.}
For each \texttt{<|univlr|>} position, the prediction is aligned with the visual target representation extracted by the frozen vision tower. To optimize both the magnitude and directional geometry of the high-dimensional latent states, we define the alignment loss $\mathcal{L}_{\mathrm{align}}$ as a combination of Layer-Normalized Mean Squared Error and Cosine Similarity:
\begin{equation}
\mathcal{L}_{\mathrm{align}} = \frac{1}{K} \sum_{k=1}^{K} \left[ \left\| \operatorname{LN}(\hat{\mathbf{z}}_k) - \operatorname{LN}(\tilde{\mathbf{z}}_k) \right\|_2^2 + \big(1 - \cos(\operatorname{LN}(\hat{\mathbf{z}}_k), \operatorname{LN}(\tilde{\mathbf{z}}_k))\big) \right]
\end{equation}
where $\operatorname{LN}(\cdot)$ denotes Layer Normalization, $\hat{\mathbf{z}}_k$ is the predicted latent token, and $\tilde{\mathbf{z}}_k$ is the offline precomputed visual target obtained via aspect-aware 2D average pooling, denoted as \texttt{pool\_avg} in our codebase. The total loss is defined as $\mathcal{L}_{\mathrm{total}} = \mathcal{L}_{\mathrm{CE}} + \lambda_{\mathrm{lvr}} \mathcal{L}_{\mathrm{align}}$, where $\lambda_{\mathrm{lvr}}$ is set to $0.1$.

\paragraph{Latent Teacher Forcing.}
To prevent error compounding during the early stages of autoregressive latent generation, we enforce full \textit{Latent Teacher Forcing} across the standard curriculum. Specifically, during the forward pass, the input embeddings corresponding to the \texttt{<|univlr|>} tokens are directly replaced by the ground-truth visual targets $\tilde{\mathbf{Z}}$. Consequently, the autoregressive context used to predict the $(k+1)$-th latent state explicitly conditions on the perfect $k$-th visual target, ensuring training stability.

\paragraph{Hyperparameters}
\label{app:hyperparameters}

All experiments are conducted using DeepSpeed ZeRO-3 optimization on 4 NVIDIA A100 GPUs. We adopt Flash Attention 2 to accelerate training. The total batch size is strictly maintained at 64 via gradient accumulation with a per-device batch size of 1. The complete set of SFT training hyperparameters is summarized in Table~\ref{tab:sft_hyperparameters}.

\begin{table}[htbp]
\centering
\caption{\textbf{Hyperparameters for UniVLR SFT Training (Stage I and Stage II).}}
\label{tab:sft_hyperparameters}
\renewcommand{\arraystretch}{1.1}
\begin{tabular}{lc}
\toprule
\textbf{Hyperparameter} & \textbf{Value} \\
\midrule
LLM Backbone & Qwen2.5-VL-7B-Instruct \\
Precision & \texttt{bfloat16} \\
Global Batch Size & 64 \\
Per-Device Batch Size & 1 \\
Optimizer & AdamW \\
Learning Rate (LLM Backbone) & $1 \times 10^{-5}$ \\
Learning Rate (UniVLR Head) & $1 \times 10^{-4}$ \\
Learning Rate Scheduler & Cosine \\
Warmup Ratio & 0.05 \\
Weight Decay & 0.1 \\
Max Gradient Norm & 1.0 \\
Epochs & 1 (per stage) \\
Training latent budget $K_{train}$ & 24 \\
Inference latent budget  $K_{infer}$ & 12 \\
Loss Coefficient $\lambda_{\mathrm{lvr}}$ & 0.1 \\
Alignment Target Layer & 14 \\
Image Token Max Pixels & $5120 \times 28 \times 28$ \\
Image Token Min Pixels & $128 \times 28 \times 28$ \\
DeepSpeed Stage & ZeRO-3 \\
\bottomrule
\end{tabular}
\end{table}
\subsection{Detailed Evaluation Setup}
\label{app:detailed_evaluation_setup}
We use VLMEvalKit \citep{Duan2024VLMEvalKit} for all our benchmark evaluations. To accommodate the unique inference mechanism of our model, we implement a customized model wrapper that extends the standard evaluation pipeline with specific decoding support for our unified visual latent tokens.

\paragraph{Benchmarks.} 
We evaluate our models on a diverse set of perception and reasoning tasks. Specifically, we evaluate guided visual search and spatial reasoning on \textbf{V*} \citep{V*}, high-resolution fine-grained perception on \textbf{HRBench4K} and \textbf{HRBench8K} \citep{HRBench}, and real-world multimodal comprehension on \textbf{MME-RealWorld-Lite} \citep{MME}. We strictly follow the evaluation protocols and data splits defined by each respective benchmark.

\paragraph{Decoding Strategy.} 
During evaluation, the model is configured to use our \texttt{UniVLR} decoding strategy. The decoding process operates deterministically i.e. greedy decoding without relying on explicit sampling parameters such as \texttt{temperature} or \texttt{top\_p}, unless overridden by specific generation configurations. 

Crucially, the decoder enforces a structured transition between the latent and textual spaces. The model first generates a predefined number of continuous visual latent blocks, bounded by the \texttt{<|univlr\_start|>} and \texttt{<|univlr\_end|>} control tokens. Upon completing the allocated latent budget, the model emits the \texttt{<|univlr\_latent\_end|>} token and seamlessly switches to standard autoregressive text generation to produce the final natural-language answer. 

In our main experiments, we use $K_{\text{infer}}=12$ continuous latent tokens for inference, while the model is trained with $K_{\text{train}}=24$ teacher-forced latent targets. In implementation, the inference wrapper truncates the latent generation phase to the first 12 latent positions before switching to answer decoding. To ensure fair evaluation by the external judges, we employ a post-processing step \texttt{clean\_univlr\_output=True} that strips all internal latent markers and unreadable token placeholders, outputting only the final textual answer.

\paragraph{System Prompts.} 
Unlike many existing models that require heavily engineered instructions, we evaluate our model in a zero-shot manner without inserting any explicit system prompts e.g. "You are a helpful visual reasoning assistant.". The input to the model consists purely of the standard dataset prompt interleaved with the visual input, formatted automatically by the base model's chat template.

\paragraph{Judging Protocol.} 
Following recent standard practices in evaluating multimodal reasoning, we adapt an LLM-as-a-Judge pipeline for performance assessment. For the V*, HRBench4K, and HRBench8K benchmarks, we employ  \texttt{gpt-4o-2024-08-06} as the external judge to assess the correctness of the generated textual answers against the ground truth. For MME-RealWorld-Lite, we utilize the standard rule-based exact-match judging protocol provided by the benchmark.
\subsection{Rendering Strategy Algorithm}
\label{app:rendering_algo}

To robustly convert multi-step explicit reasoning traces and multi-modal auxiliary images into unified visual canvases, we propose three distinct rendering algorithms. Unlike naive fixed-size rendering which may cause severe text truncation or excessive blank areas, our strategies dynamically adjust the canvas dimensions and component layouts based on the content. The specific strategy applied depends on the reasoning step pattern and the dataset construction stage.

Table~\ref{tab:rendering_strategies_overview} summarizes the three rendering strategies and their core parameter configurations.

\vspace{10pt} 
\noindent 
\begin{minipage}{\textwidth}
\centering
\captionof{table}{\textbf{Overview of MCoT Canvas Rendering Strategies and Configurations.}}
\label{tab:rendering_strategies_overview}
\renewcommand{\arraystretch}{1.15}
\resizebox{\textwidth}{!}{
\begin{tabular}{lp{4cm}p{4cm}p{5.5cm}}
\toprule
\textbf{Strategy} & \textbf{Key Characteristics} & \textbf{Applicable Patterns} & \textbf{Core Hyperparameters} \\
\midrule
\textbf{Vertical Layout} (Algorithm \ref{alg:vertical}) 
& Dynamic height, sequential top-to-bottom cards.
& Long sequential reasoning traces. 
& \texttt{min\_canvas\_width}: 420px, \newline \texttt{outer\_padding}: 24px, \texttt{gap}: 18px. \\

\textbf{Compact Left-Right Layout} (Algorithm \ref{alg:compact_lr}) 
& Dynamic height, left image panel, right text cards with directed flow arrows.
& \texttt{(t,j,t)} or \texttt{(t,j,t,t)} patterns with complex spatial focus.
& \texttt{canvas\_width}: 1536px, \newline \texttt{max\_body\_height}: 1180px, \newline \texttt{column\_gap}: 28px. \\

\textbf{Fixed-Canvas Adaptive Wrap} (Algorithm \ref{alg:fixed_wrap}) 
& Fixed $W \times H$ canvas, bottom-left image, adaptive font-size text wrapping.
& General visual grounding and baseline comparisons. 
& \texttt{canvas\_width}: 1024px, \newline \texttt{canvas\_height}: 1024px, \newline \texttt{font\_range}: $[14, 80]$px. \\
\bottomrule
\end{tabular}
}
\end{minipage}
\vspace{5pt} 

\paragraph{1. Vertical Layout Strategy.}
This strategy is utilized to render lengthy reasoning traces where components naturally follow a sequential logic. As outlined in Algorithm \ref{alg:vertical}, the canvas width is constrained by a minimum threshold and the scaled auxiliary image. Each textual or visual reasoning block is rendered into a rounded card, and the final canvas height is dynamically computed to accommodate all vertically stacked cards. Connecting arrows are drawn between successive cards to indicate logical progression.

\begin{algorithm}[htbp]
\caption{Vertical Layout Strategy}
\label{alg:vertical}
\textbf{Input:} Ordered reasoning blocks $\mathcal{B} = \{b_1, \dots, b_N\}$, Auxiliary image $I$, min\_width $W_{\min}$, padding $p$, gap $g$. \\
\textbf{Output:} Unified Canvas $C_{\text{vert}}$.
\begin{algorithmic}[1]
\STATE \textbf{Determine Canvas Width:}
\STATE Highlight missing/focus regions on $I$ to obtain $I'$.
\STATE $W_{\text{canvas}} \leftarrow \max(W_{\min}, I'_{width} + 2 \times p)$
\STATE \textbf{Construct Component Cards:}
\STATE Initialize card list $\mathcal{L}_{\text{cards}} \leftarrow[]$
\FOR{each block $b \in \mathcal{B}$}
    \IF{$b$ is a joint image block}
        \STATE Render card $C_{b}$ containing $I'$ and block text, constrained to $W_{\text{canvas}} - 2p$.
    \ELSE
        \STATE Render textual card $C_{b}$ containing block text, constrained to $W_{\text{canvas}} - 2p$.
    \ENDIF
    \STATE Append $C_{b}$ to $\mathcal{L}_{\text{cards}}$.
\ENDFOR
\STATE \textbf{Dynamic Canvas Assembly:}
\STATE $H_{\text{canvas}} \leftarrow 2p + \sum_{C_b \in \mathcal{L}_{\text{cards}}} \text{height}(C_b) + g \times (N - 1)$
\STATE Initialize blank canvas $C_{\text{vert}}$ with size $(W_{\text{canvas}}, H_{\text{canvas}})$.
\STATE $Y_{cursor} \leftarrow p$
\FOR{$i = 1$ \TO $N$}
    \STATE Paste $\mathcal{L}_{\text{cards}}[i]$ onto $C_{\text{vert}}$ at $(p, Y_{cursor})$.
    \IF{$i < N$}
        \STATE Draw a vertical directed arrow from the bottom of $\mathcal{L}_{\text{cards}}[i]$ to the top of $\mathcal{L}_{\text{cards}}[i+1]$.
    \ENDIF
    \STATE $Y_{cursor} \leftarrow Y_{cursor} + \text{height}(\mathcal{L}_{\text{cards}}[i]) + g$
\ENDFOR
\STATE \textbf{Return} $C_{\text{vert}}$.
\end{algorithmic}
\end{algorithm}

\paragraph{2. Compact Left-Right Layout Strategy.}
For reasoning patterns involving a central visual grounding step surrounded by textual analysis (e.g., text $\rightarrow$ joint-image $\rightarrow$ text), we employ a compact left-right architecture. As detailed in Algorithm \ref{alg:compact_lr}, the canvas has a fixed width but dynamic height. The left panel is dedicated to the highlighted auxiliary image and its localized description, while the right panel vertically stacks the pure-text reasoning cards. Directional arrows are rendered across columns to trace the multimodal logical dependencies explicitly.

\begin{algorithm}[htbp]
\caption{Compact Left-Right Layout Strategy}
\label{alg:compact_lr}
\textbf{Input:} Text blocks $\mathcal{B}_{\text{text}}$, Joint block $b_{\text{joint}}$, Auxiliary image $I$, canvas\_width $W$, max\_body\_height $H_{\max}$, margin $m$, column\_gap $g_{\text{col}}$. \\
\textbf{Output:} Unified Canvas $C_{\text{compact}}$.
\begin{algorithmic}[1]
\STATE \textbf{Column Partitioning:}
\STATE $W_{\text{left}} \leftarrow (W - 2m - g_{\text{col}}) / 2$
\STATE $W_{\text{right}} \leftarrow W - 2m - g_{\text{col}} - W_{\text{left}}$
\STATE \textbf{Left Panel Construction:}
\STATE Highlight focus regions on $I$ to obtain $I'$.
\STATE Resize $I'$ to fit within $W_{\text{left}}$ and $H_{\max}$.
\STATE Calculate joint text height $H_{\text{j-text}}$ and assemble the left joint panel $P_{\text{left}}$.
\STATE \textbf{Right Panel Construction \& Height Synchronization:}
\STATE Calculate base height $H_{\text{body}} \leftarrow \text{height}(P_{\text{left}}) + \text{offset}$
\STATE Determine uniform row height $H_{\text{row}}$ for right-side text cards based on $H_{\text{body}}$ and number of text blocks $|\mathcal{B}_{\text{text}}|$.
\STATE Recalculate final $H_{\text{canvas}}$ to perfectly align both columns.
\STATE Initialize blank canvas $C_{\text{compact}}$ with size $(W, H_{\text{canvas}})$.
\STATE \textbf{Assembly and Dependency Tracking:}
\STATE Paste $P_{\text{left}}$ at $(m, m)$. Record its bounding box.
\STATE $Y_{cursor} \leftarrow m$
\FOR{each block $b_t \in \mathcal{B}_{\text{text}}$}
    \STATE Render textual card $C_t$ with size $(W_{\text{right}}, H_{\text{row}})$.
    \STATE Paste $C_t$ at $(m + W_{\text{left}} + g_{\text{col}}, Y_{cursor})$. Record bounding box.
    \STATE $Y_{cursor} \leftarrow Y_{cursor} + H_{\text{row}} + \text{row\_gap}$
\ENDFOR
\STATE Draw horizontal and vertical arrows between recorded bounding boxes based on the original chronological thought sequence 
\STATE \textbf{Return} $C_{\text{compact}}$.
\end{algorithmic}
\end{algorithm}

\paragraph{3. Fixed-Canvas Adaptive Wrap Strategy.}
To ensure strict resolution control during certain experimental settings, we design a fixed-canvas strategy. As described in Algorithm \ref{alg:fixed_wrap}, the canvas dimensions are rigidly constrained (e.g., $1024 \times 1024$). The auxiliary image is scaled and anchored to the bottom-left corner. The textual reasoning is then rendered using an adaptive font-size search mechanism. The text wraps dynamically: lines above the image utilize the full canvas width, while lines adjacent to the image wrap exclusively within the remaining right-side space, perfectly avoiding the image boundaries.

\begin{algorithm}[htbp]
\caption{Fixed-Canvas Adaptive Wrap Strategy}
\label{alg:fixed_wrap}
\textbf{Input:} Textual reasoning trace $\mathcal{T}$, Auxiliary image $I$, Fixed canvas size $(W, H)$, padding $p$, gap $g$. \\
\textbf{Output:} Unified Canvas $C_{\text{fixed}}$.
\begin{algorithmic}[1]
\STATE \textbf{Image Anchor Placement:}
\STATE Scale $I$ such that it occupies at most 50\% of canvas width and height.
\STATE Place $I$ at the bottom-left corner: $X_{img} \leftarrow p$, $Y_{img} \leftarrow H - p - I_{height}$.
\STATE \textbf{Adaptive Font Search \& Text Wrapping:}
\STATE Initialize optimal font $f_{\text{opt}} \leftarrow \text{None}$.
\FOR{font size $f \in \{f_{\max}, f_{\max}-1, \dots, f_{\min}\}$}
    \STATE $Y_{cursor} \leftarrow p$
    \STATE Initialize current line text.
    \FOR{each word $w \in \mathcal{T}$}
        \STATE \textbf{Dynamic Width Constraint:}
        \IF{$Y_{cursor} \ge Y_{img}$ \AND $Y_{cursor} \le Y_{img} + I_{height}$}
            \STATE $W_{\text{avail}} \leftarrow W - (X_{img} + I_{width} + g) - p$
        \ELSE
            \STATE $W_{\text{avail}} \leftarrow W - 2p$
        \ENDIF
        \STATE Test if appending $w$ exceeds $W_{\text{avail}}$. If yes, commit line and advance $Y_{cursor} \leftarrow Y_{cursor} + f + (f/4)$.
    \ENDFOR
    \IF{$Y_{cursor} \le H - p$}
        \STATE $f_{\text{opt}} \leftarrow f$ \COMMENT{Found the largest font that fits all text.}
        \STATE \textbf{Break}
    \ENDIF
\ENDFOR
\STATE If $f_{\text{opt}}$ is not found, fallback to $f_{\min}$ and truncate overflowing text.
\STATE Initialize blank canvas $C_{\text{fixed}}$ with size $(W, H)$.
\STATE Paste $I$ at $(X_{img}, Y_{img})$ and draw text $\mathcal{T}$ using the computed layout boundaries.
\STATE \textbf{Return} $C_{\text{fixed}}$.
\end{algorithmic}
\end{algorithm}
\subsection{Rendering Strategy Examples}
\label{app:rendering_examples}

To provide an intuitive understanding of our adaptive rendering algorithms, we present qualitative examples of the unified MCoT canvases generated by each of the three strategies. As demonstrated below, our rendering module effectively accommodates diverse auxiliary image aspect ratios and textual reasoning lengths, ensuring that the resulting visual supervision targets are structurally coherent and semantically dense.

\begin{figure}[htbp]
    \centering
    \includegraphics[width=0.85\linewidth]{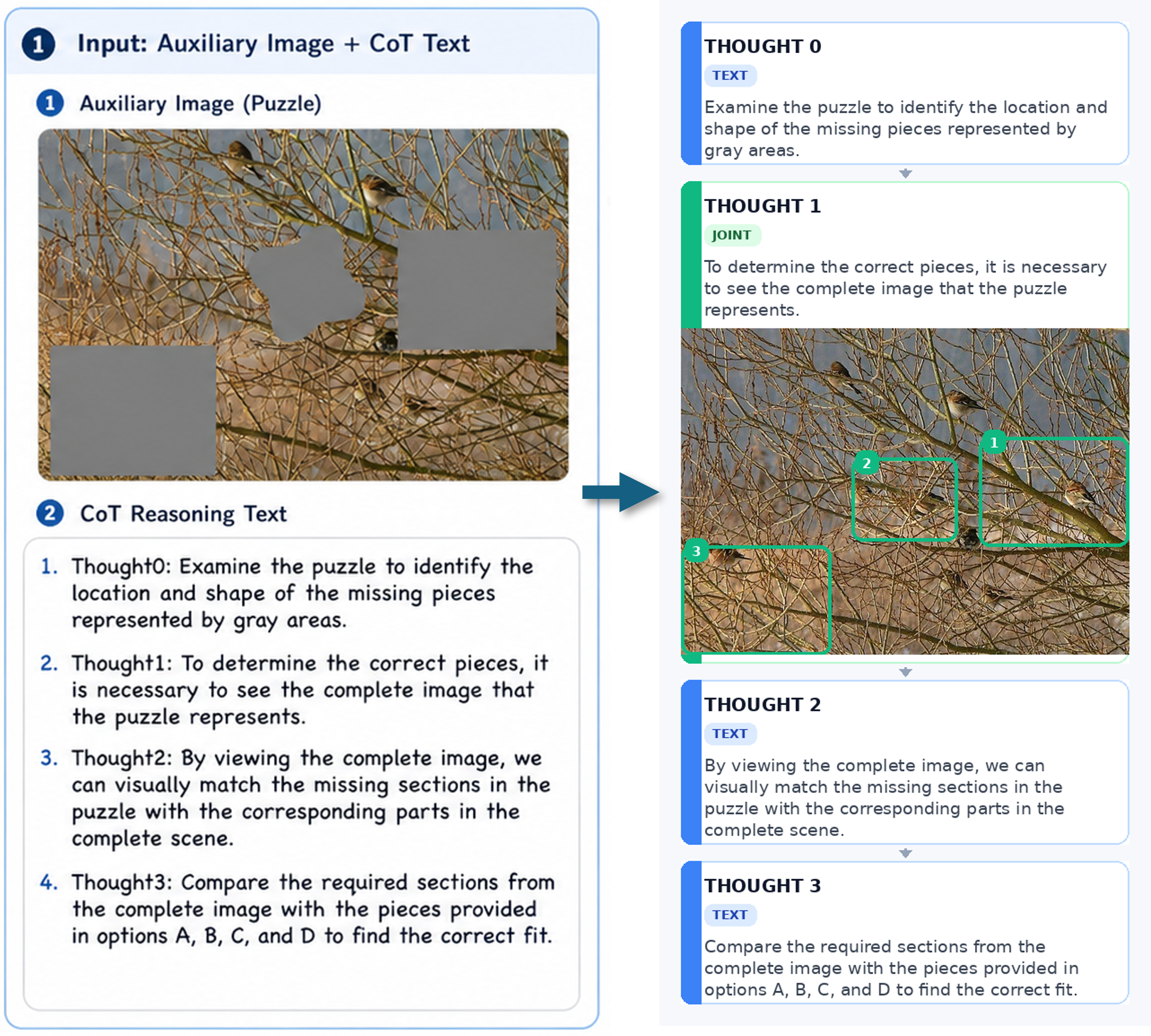} 
    \vspace{2pt}
    \caption{
    \textbf{Example of the Vertical Layout Strategy.} Generated by Algorithm \ref{alg:vertical}, this layout dynamically extends the canvas height to accommodate a long sequence of reasoning steps.
    }
    \label{fig:render_vertical}
\end{figure}

\begin{figure}[htbp]
    \centering

    \includegraphics[width=0.85\linewidth]{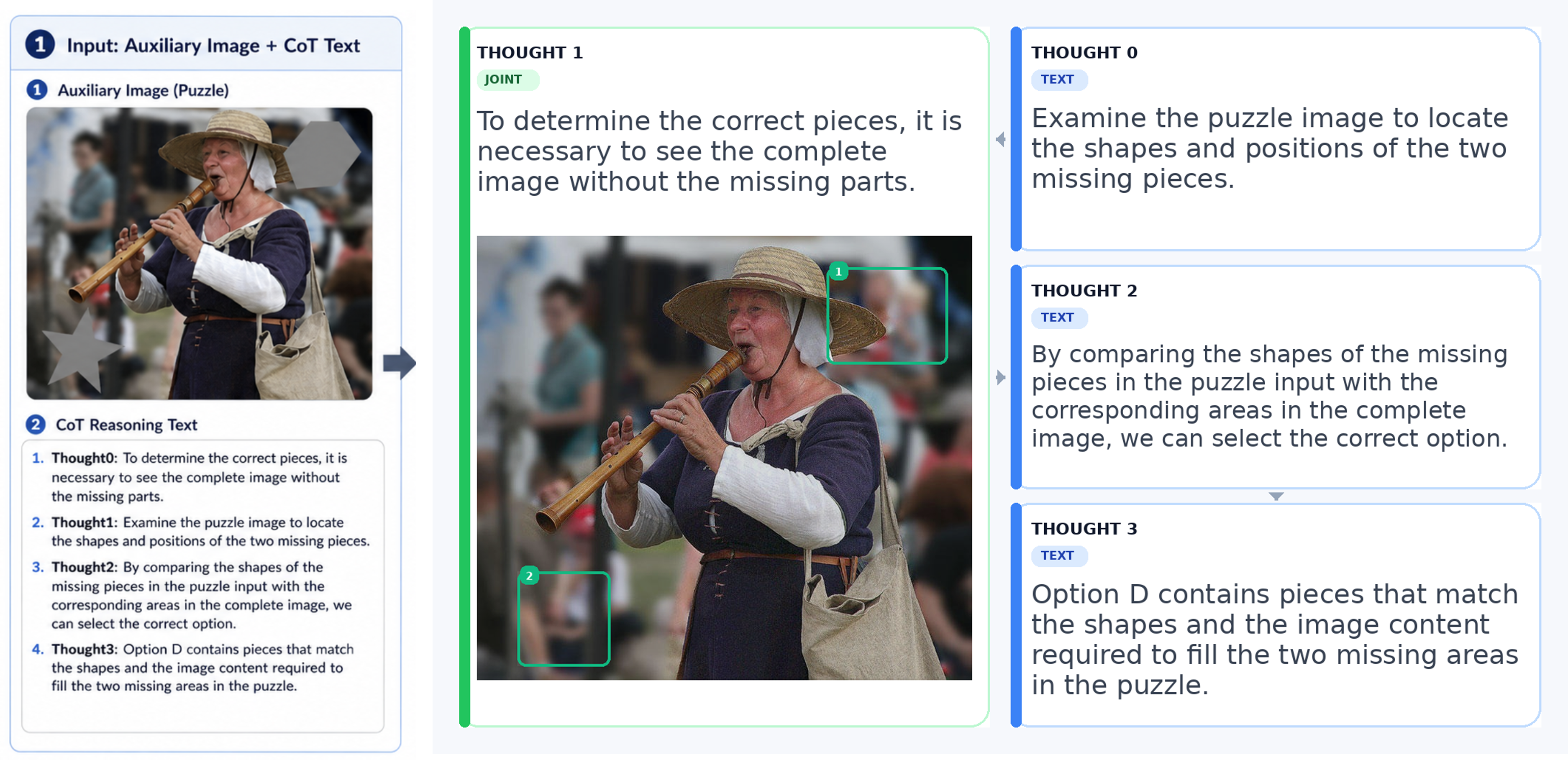} 
    \vspace{2pt}
    \caption{
    \textbf{Example of the Compact Left-Right Layout Strategy.} Generated by Algorithm \ref{alg:compact_lr}, this layout partitions the canvas into two columns. The left panel houses the highlighted auxiliary image (serving as the visual grounding anchor), while the right panel vertically stacks the text-only reasoning cards. Directional flow arrows bridge the two columns, visually encoding the multimodal dependencies (e.g., text $\rightarrow$ joint $\rightarrow$ text).
    }
    \label{fig:render_compact_lr}
\end{figure}

\begin{figure}[htbp]
    \centering

    \includegraphics[width=0.85\linewidth]{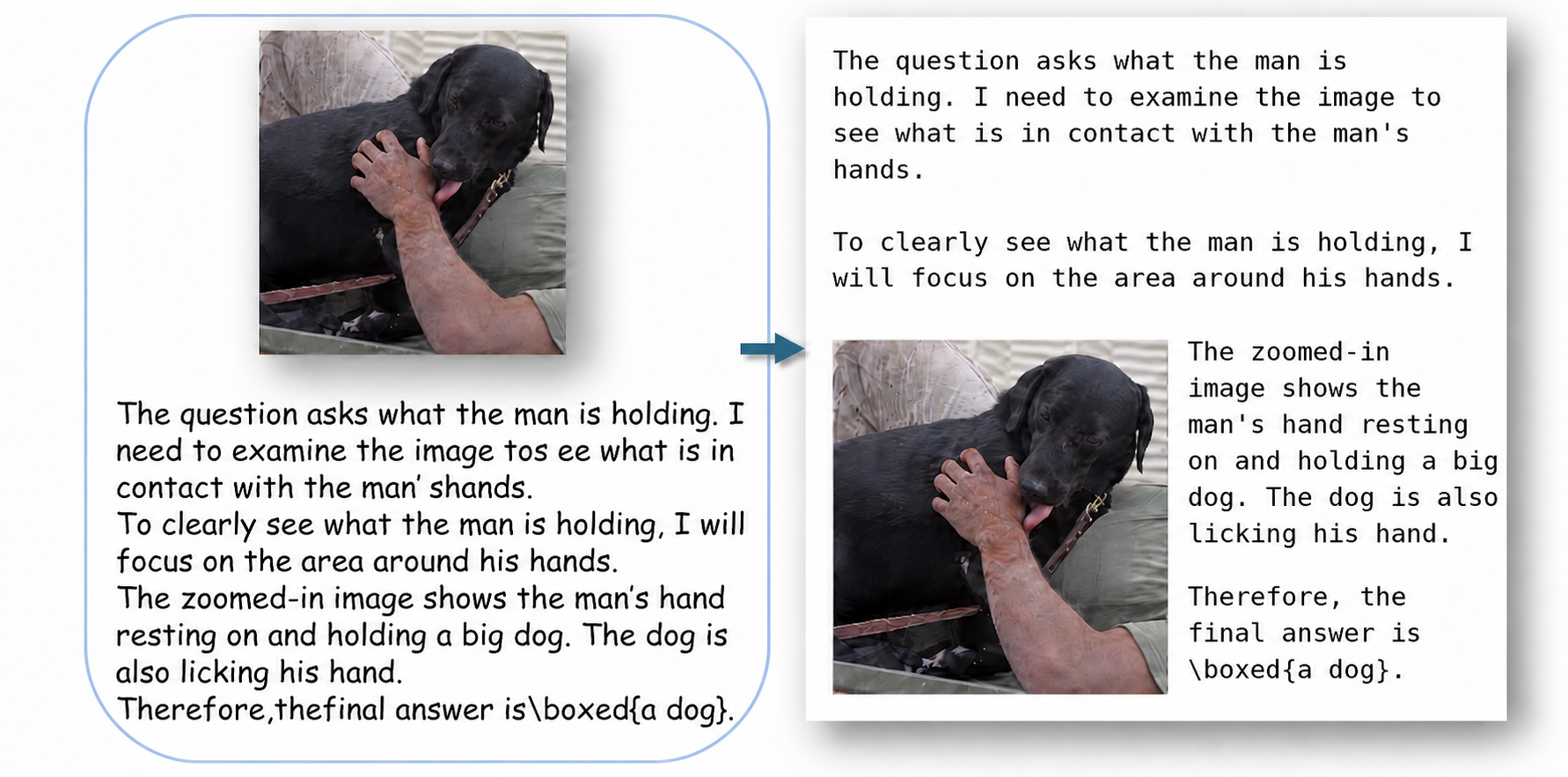} 
    \vspace{2pt}
    \caption{
    \textbf{Example of the Fixed-Canvas Adaptive Wrap Strategy.} Generated by Algorithm \ref{alg:fixed_wrap}, the canvas dimensions are strictly bounded $1024 \times 1024$. 
    }
    \label{fig:render_fixed_wrap}
\end{figure}
\section{UniVLR-SFT-140K Construction}
\label{app:dataset_construction}

A core contribution of our work is constructing high-quality, dense visual supervision signals to train the latent reasoning interface. The unfiltered data may introduce shortcut-prone supervision, weakening the visual-latent mapping established in Stage I. By retaining 21.5K filtered Zebra-CoT samples, we reduce noisy supervision and encourage a stronger dependency between the unified canvas targets and final answers.

\subsection{Data Statistics}
\label{app:data_statistics}

Our UniVLR-SFT-140K dataset is primarily curated from the Visual-CoT subset of Monet-SFT-125K~\cite{monet} and highly structured subsets of Zebra-CoT~\cite{li2025zebra}. To effectively implement our two-stage curriculum, we utilize different data compositions in each stage:
\begin{itemize}[leftmargin=*]
    \item \textbf{Stage I (Latent Warm-up):} We utilize the entire Visual-CoT dataset, comprising 118K samples, as the foundational training corpus. This dataset, featuring abstract visual operations such as bounding boxes, serves to establish the initial mapping between the continuous latent space and the structural topology of the unified visual canvases.
    \item \textbf{Stage II (Curriculum Mixing):} We curate a high-quality mixed dataset containing approximately 30.7K instances. This mixture consists of all 21.5K rigorously filtered samples from three Zebra-CoT subsets (Visual Search, Jigsaw, and Maze) and a randomly sampled portion of the Visual-CoT dataset, maintaining a 7:3 ratio between the complex Zebra-CoT data and the foundational Visual-CoT data.
\end{itemize}

The detailed statistics, problem domains, and visual operation types of our final UniVLR-SFT-140K dataset are summarized in Table~\ref{tab:data_statistics}.

\begin{table}[htbp]
\centering
\caption{\textbf{Statistics of the UniVLR-SFT-140K.} The dataset supports our two-stage curriculum, utilizing the full Visual-CoT for Stage I and a 7:3 mixture of heavily filtered Zebra-CoT subsets and sampled Visual-CoT for Stage II.}
\label{tab:data_statistics}
\renewcommand{\arraystretch}{1.15}
\resizebox{\textwidth}{!}{
\begin{tabular}{lllc}
\toprule
\textbf{Data Source} & \textbf{Problem Domain} & \textbf{Visual Operation Type} & \textbf{Amount} \\
\midrule
Visual-CoT & Real-world, Documents, Charts & Abstract visual operations (e.g., bounding boxes) & 118.6K \\
Zebra-CoT Visual Search & 2D Visual Reasoning & Object localization, local focus highlighting & 8.7K \\
Zebra-CoT Jigsaw & 2D Visual Reasoning & Spatial geometry, piece alignment & 8.8K \\
Zebra-CoT Maze & Visual Logic \& Strategic Games & Pathfinding, spatial constraint satisfaction & 4.0K \\
\midrule
\multicolumn{3}{l}{\textbf{Total (UniVLR-SFT-140K)}} & \textbf{140.1K} \\
\bottomrule
\end{tabular}
}
\end{table}

\paragraph{Comparison with Existing Latent Reasoning Datasets.}
To contextualize our data curation efforts, we compare the scale and composition of our training corpora with recent state-of-the-art latent and visual reasoning models.
\begin{itemize}[leftmargin=*]
    \item \textbf{Monet}~\cite{monet} utilizes Monet-SFT-125K, which predominantly consists of Visual-CoT alongside smaller portions of ReFocus, CogCoM, and Zebra-CoT.
    \item \textbf{CoVT}~\cite{COVT} constructs a substantial dataset derived primarily from LLaVA-OneVision vision-centric subsets and a filtered TallyQA counting dataset.
    \item \textbf{LVR}~\cite{LVR} employs the full Visual-CoT dataset (438K instances) for its supervised fine-tuning stage.
    \item \textbf{SkiLa}~\cite{skila} curates a 101K sample dataset entirely filtered from Zebra-CoT, excluding 3D data and excessively complex sketch images.
\end{itemize}
Our UniVLR-SFT-140K distinctively organizes its 140K samples into a two-stage curriculum, prioritizing the quality and structural regularity of the visual supervision signals over raw scale.

\subsection{Data Filtering}
\label{app:data_filtering}

To ensure that the latent tokens actively learn to encode essential visual semantics rather than memorizing textual shortcuts, and to maintain the high quality of the visual targets, we design a rigorous filtering pipeline consisting of bidirectional model-based filtering and geometric layout filtering.

\paragraph{Step 1: Lower-Bound Filtering.}
Many instances in standard multimodal reasoning datasets appear complex but can actually be answered correctly using pure linguistic priors or shallow visual scanning, without needing the intermediate visual reasoning steps. Training on these instances encourages the latent tokens to bypass genuine visual perception. To eliminate this, we feed the raw \textit{Question} and \textit{Problem Image} without any auxiliary reasoning canvas into \textbf{Qwen2.5-VL-7B}. If the model correctly answers the question zero-shot, we discard the sample, retaining only those that strictly require intermediate visual reasoning to solve.

\paragraph{Step 2: Upper-Bound Filtering.}
Conversely, some reasoning traces contain severe logical flaws or mismatched auxiliary images, which corrupt the latent alignment target. To filter these out, we feed the \textit{Question}, \textit{Problem Image}, and the \textit{Unified MCoT Canvas} or auxiliary images into a much stronger teacher model, \textbf{Qwen2.5-VL-72B}. If this powerful teacher still fails to derive the correct answer despite having access to the explicit visual reasoning steps, we deem the sample unsolvable or noisy and discard it.

\paragraph{Step 3: Aspect-Ratio Filtering for Canvas Density.}
Besides semantic correctness, the geometric properties of the auxiliary images significantly impact the quality of our unified visual representations. Auxiliary images with extreme aspect ratios i.e., excessively long or wide force our rendering algorithm to apply aggressive scaling and padding to fit within the canvas boundaries. This inevitably introduces massive whitespace regions onto the unified canvas, which dilutes the information density of the extracted visual features and wastes the limited capacity of the continuous latent tokens. Therefore, we filter out samples containing auxiliary images with extreme aspect ratios, ensuring that the rendered MCoT canvas remains compact, semantically dense, and visually balanced for the vision encoder.

\paragraph{Impact of Filtering on Latent Alignment.}
Our empirical observations as detailed in Section \ref{exp:ablation} confirm that this bidirectional filtering is critical for stable Stage II training. When we attempted to mix the raw, unfiltered Zebra-CoT data $\sim$40K instances with Visual-CoT, the downstream accuracy degraded. The unfiltered data diluted the strong visual-latent mapping established in Stage I because the fake hard samples taught the model to exploit spurious correlations, causing the latent representation to collapse. By retaining only the rigorously filtered $21.5K$ Zebra-CoT samples, we enforce a strict causal dependency between the visual canvas target and the final answer, significantly improving the efficacy of the unified visual latent reasoning.

\section{Additional Experimental Results}
\label{app:additional_experiment_results}

In this section, we provide additional experimental results to complement the analyses in the main paper. 
These experiments further examine the sensitivity of UniVLR to the latent-alignment loss weight, its generalization behavior beyond the core visual reasoning benchmarks, and the effect of latent teacher forcing during continuous autoregressive training. 
Unless otherwise specified, we follow the same evaluation protocol as in the main text and report results using the corresponding UniVLR checkpoint described in each subsection.

\subsection{Ablation Study for \texorpdfstring{$\lambda_{\mathrm{align}}$}{lambda\_align}}
\label{app:lambda_ablation}

UniVLR is trained with a combination of the standard autoregressive language modeling objective and the latent representation alignment objective. 
The coefficient $\lambda_{\mathrm{align}}$ controls the relative strength of the visual latent supervision. 
A very small value may provide insufficient alignment to the visual representation space, whereas an overly large value may over-constrain the latent states and reduce their flexibility for downstream answer prediction. 
We therefore vary $\lambda_{\mathrm{align}} \in \{0.1, 0.3, 0.5, 0.7\}$ and report the results in Table~\ref{tab:lambda_ablation}.

Overall, $\lambda_{\mathrm{align}}=0.1$ provides the best accuracy--stability trade-off across the evaluated benchmarks. 
Increasing the alignment weight generally leads to performance degradation, especially on MME-RealWorld-Lite. 
This suggests that, while visual latent alignment is important, excessive alignment pressure can make the latent trajectory less adaptive to task-specific reasoning and answer decoding.

\begin{table}[htbp]
\centering
\caption{\textbf{Ablation study for \texorpdfstring{$\lambda_{\mathrm{align}}$}{lambda\_align}.} 
We vary the weight of the latent alignment objective while keeping the remaining training and evaluation settings unchanged.}
\label{tab:lambda_ablation}
\renewcommand{\arraystretch}{1.1}
\setlength{\tabcolsep}{3.5pt}
\begin{adjustbox}{max width=\textwidth}
\begin{tabular}{l ccc ccc ccc ccc}
\toprule
\multirow{2.5}{*}{\textbf{Model}} & \multicolumn{3}{c}{\textbf{V*}} & \multicolumn{3}{c}{\textbf{HRBench4K}} & \multicolumn{3}{c}{\textbf{HRBench8K}} & \multicolumn{3}{c}{\textbf{MME-RealWorld-Lite}} \\
\cmidrule(lr){2-4} \cmidrule(lr){5-7} \cmidrule(lr){8-10} \cmidrule(lr){11-13}
& Overall & Attr. & Spa. & Overall & FSP & FCP & Overall & FSP & FCP & Overall & Rea. & Perc. \\
\midrule
$\lambda_{\text{align}}=0.1$ & \textbf{82.7} & \textbf{83.5} & \textbf{81.6} 
        & \textbf{73.3} & \textbf{86.0} & \textbf{60.5}
        & \textbf{68.8} & \textbf{78.8} & \textbf{58.8} 
        & \textbf{50.7} & \textbf{44.7} & \textbf{54.5} \\
$\lambda_{\text{align}}=0.3$ & 82.7 & 83.5 & 81.6 & 72.8 & 86.0 & 59.5 & 66.6 & 75.5 & 57.8 & 46.9 & 42.0 & 50.1 \\
$\lambda_{\text{align}}=0.5$ & 81.7 & 83.5 & 78.9 & 72.5 & 85.5 & 59.5 & 65.3 & 75.0 & 55.5 & 25.4 & 22.4 & 27.3 \\
$\lambda_{\text{align}}=0.7$ & 82.2 & 82.6 & 81.6 & 71.8 & 85.5 & 58.0 & 65.5 & 75.3 & 55.8 & 29.0 & 28.8 & 29.2 \\
\bottomrule
\end{tabular}
\end{adjustbox}
\end{table}

\subsection{Generalization on Diverse Multimodal Tasks}
\label{app:generalization}

A potential concern when fine-tuning MLLMs for specialized latent visual reasoning is catastrophic forgetting: the model may improve on the target reasoning tasks but lose general multimodal capabilities. 
To examine this issue, we evaluate UniVLR on a broader set of benchmarks that are not exclusively designed for complex visual latent reasoning. 
Specifically, we consider \textbf{TextVQA}~\cite{singh2019towards} for text-rich image understanding, \textbf{MME$_{\text{translation}}$}~\cite{fu2023mme} for multilingual multimodal alignment, \textbf{POPE}~\cite{li2023evaluating} for object hallucination evaluation, and \textbf{WeMath$_{\text{loose}}$}~\cite{qiao2024we} for abstract mathematical reasoning.

The results are summarized in Table~\ref{tab:generalization}. 
Overall, UniVLR does not exhibit severe degradation on general-purpose multimodal tasks. 
Instead, it improves over the base model on TextVQA and POPE, while maintaining competitive performance on MME$_{\text{translation}}$ and WeMath$_{\text{loose}}$.

\begin{table*}[htbp]
    \centering
    \caption{
    \textbf{Generalization performance on diverse multimodal benchmarks.} 
    We evaluate whether UniVLR preserves general multimodal capabilities beyond the core visual reasoning tasks. 
    Best results are highlighted in \textbf{bold}.
    }
    \label{tab:generalization}
    \vspace{2pt}
    \renewcommand{\arraystretch}{1.15}
    \resizebox{\textwidth}{!}{
    \begin{tabular}{l c c cccc c}
        \toprule
        \multirow{2}{*}{\textbf{Model}} & \multirow{2}{*}{\textbf{TextVQA}} & \multirow{2}{*}{\textbf{MME$_{\text{translation}}$}} & \multicolumn{4}{c}{\textbf{POPE}} & \multirow{2}{*}{\textbf{WeMath$_{\text{loose}}$}} \\
        \cmidrule(lr){4-7}
        & & & \textbf{Overall} & Adversarial & Popular & Random & \\
        \midrule
        Qwen2.5-VL-7B~\cite{qwen2.5-vl} & 77.5 & 185.0 & 86.4 & 85.5 & 86.5 & 87.2 & \textbf{52.1} \\
        \midrule
        LVR~\cite{LVR} & 75.6 & 200.0 & 84.9 & 84.3 & 84.8 & 85.7 & 48.4 \\
        SkiLa~\cite{skila} & 79.3 & \textbf{200.0} & 87.1 & 86.2 & 87.2 & 87.9 & 46.2 \\
        CoVT~\cite{COVT} & 66.5 & 192.5 & 88.7 & 87.2 & 88.6 & 90.3 & 50.1 \\
        \midrule
        UniVLR-Stage1 & 79.1 & 183.0 & 84.2 & 83.7 & 84.2 & 84.8 & 48.6 \\
        \rowcolor{gray!20} \textbf{UniVLR (Ours)} & \textbf{80.4} & \textbf{200.0} & \textbf{88.8} & \textbf{87.3} & \textbf{88.9} & \textbf{90.4} & 49.4 \\
        \bottomrule
    \end{tabular}
    }
\end{table*}

\paragraph{Perceptual grounding and hallucination mitigation.}
UniVLR achieves the best performance on TextVQA and POPE among the compared models. 
The improvement on TextVQA suggests that the unified visual latent training does not weaken text-rich image understanding; instead, it can improve the model's ability to extract and use visual textual evidence. 
On POPE, UniVLR consistently improves over the base Qwen2.5-VL-7B model across the adversarial, popular, and random splits. 
Since POPE is designed to measure object hallucination, these gains indicate that the learned visual latent channel may encourage the model to rely more strongly on grounded visual evidence rather than language priors.

We attribute this behavior to the proposed unified canvas supervision. 
During training, UniVLR aligns latent reasoning states with dense visual representations that jointly encode textual reasoning traces and auxiliary visual evidence. 
This encourages the latent tokens to remain anchored to the visual input, which may reduce unsupported object predictions and improve recognition of text and fine-grained visual content.

\paragraph{Generalization on abstract reasoning tasks.}
UniVLR also remains competitive on tasks that are less directly tied to visual latent reasoning. 
It achieves a perfect score of 200.0 on MME$_{\text{translation}}$, indicating that the fine-tuning process does not degrade multilingual multimodal alignment in this setting. 
On WeMath$_{\text{loose}}$, UniVLR shows a moderate decrease compared with the base Qwen2.5-VL-7B model, from 52.1 to 49.4. 
This suggests a small specialization trade-off introduced by the visual reasoning curriculum. 
Nevertheless, UniVLR remains competitive with other visual latent reasoning baselines, outperforming LVR and SkiLa on this benchmark. 
These results suggest that UniVLR improves visual grounding and hallucination robustness without causing severe degradation in general multimodal capabilities.

\subsection{Ablation Study on Latent Teacher Forcing}
\label{app:teacher_forcing}

Continuous latent reasoning introduces a training--inference discrepancy. 
During training, the model can condition on target latent embeddings, whereas during inference it must recursively condition on its own predicted latent states. 
In discrete autoregressive generation, scheduled sampling~\citep{DBLP:journals/corr/BengioVJS15} is often used to reduce such exposure bias by gradually replacing ground-truth inputs with model predictions. 
We investigate whether this strategy is also beneficial for continuous visual latent generation.

We compare full latent teacher forcing with a scheduled sampling variant during Stage II training. 
In the full teacher-forcing setting, all $K=24$ latent inputs are replaced with the corresponding ground-truth visual targets. 
In the scheduled sampling variant, teacher forcing is applied only to the first 12 latent slots, while the remaining 12 slots use the model's detached predictions as inputs, which we refer to as half-replay. 
The results are reported in Table~\ref{tab:scheduled_sampling}.

\begin{table}[htbp]
\centering
\caption{\textbf{Effect of latent teacher forcing and scheduled sampling.} 
Experiments are conducted on the Stage II model. 
Full latent teacher forcing yields more stable training and better downstream accuracy across most benchmark metrics.}
\label{tab:scheduled_sampling}
\renewcommand{\arraystretch}{1.15}
\resizebox{\textwidth}{!}{
\begin{tabular}{l ccc ccc ccc ccc}
\toprule
\multirow{2.5}{*}{\textbf{Training Strategy}} & \multicolumn{3}{c}{\textbf{V*}} & \multicolumn{3}{c}{\textbf{HRBench4K}} & \multicolumn{3}{c}{\textbf{HRBench8K}} & \multicolumn{3}{c}{\textbf{MME-RealWorld-Lite}} \\
\cmidrule(lr){2-4} \cmidrule(lr){5-7} \cmidrule(lr){8-10} \cmidrule(lr){11-13}
& Overall & Attr. & Spa. & Overall & FSP & FCP & Overall & FSP & FCP & Overall & Rea. & Perc. \\
\midrule
\rowcolor{gray!20} \textbf{Full Latent Teacher Forcing} & \textbf{82.7} & \textbf{83.5} & \textbf{81.6} & \textbf{73.3} & \textbf{86.0} & \textbf{60.5} & \textbf{68.8} & \textbf{78.8} & \textbf{58.8} & \textbf{50.7} & 44.7 & \textbf{54.5} \\
Scheduled Sampling (Half-Replay) & 82.2 & 82.6 & \textbf{81.6} & 72.4 & 85.0 & 59.8 & 66.8 & 76.5 & 57.0 & 48.8 & \textbf{44.9} & 51.2 \\
\bottomrule
\end{tabular}
}
\end{table}

\paragraph{Analysis.}
Contrary to the common intuition from discrete text generation, scheduled sampling does not improve performance in our continuous latent reasoning setting. 
Compared with full latent teacher forcing, half-replay reduces the overall score by 2.0 points on HRBench8K and 1.9 points on MME-RealWorld-Lite. 
It also underperforms on most fine-grained subcategories, with the only exception being a small gain on the reasoning split of MME-RealWorld-Lite.

We hypothesize that this behavior arises from the geometry of high-dimensional continuous latent spaces. 
In early training, the model's predicted latent vectors may not yet lie on a stable visual representation manifold. 
Feeding these imperfect continuous predictions back into the model can therefore inject semantic noise into the autoregressive context, which in turn affects subsequent latent predictions and weakens the alignment objective. 
By contrast, full latent teacher forcing provides a cleaner and more stable supervision signal at every latent step, making representation learning easier and leading to stronger downstream performance. 
This result suggests that, for visual latent reasoning, stabilizing the latent manifold during training can be more important than directly reducing exposure bias through self-replay.

\subsection{Additional Efficiency Evaluation}
\label{app:additional_efficiency}

We provide an additional efficiency evaluation on two representative benchmarks, HRBench8K and MME-RealWorld-Lite. 
The goal of this experiment is to complement the reasoning-token analysis in the main paper with a lightweight runtime and memory measurement under the same evaluation pipeline.

We note that output-token throughput is not an ideal metric for this setting. 
Visual latent reasoning methods often generate very short final textual answers, and UniVLR uses only a compact latent reasoning budget before answer decoding. 
Consequently, output tokens per second can be dominated by fixed evaluation overheads and small variations in answer length. 
We therefore report average per-sample time, total sample time, peak allocated GPU memory, and peak reserved GPU memory. 
All methods are evaluated with the same benchmark wrapper and logging protocol.

\begin{table}[htbp]
\centering
\caption{
\textbf{Additional efficiency comparison on HRBench8K and MME-RealWorld-Lite.}
We report average per-sample time, total sample time, and peak allocated/reserved GPU memory. 
All values are rounded to one decimal place. 
The relative reductions over the Qwen2.5-VL-7B baseline are shown with {\color{teal}\textbf{$\downarrow$}}.
}
\label{tab:additional_efficiency}
\renewcommand{\arraystretch}{1.15}
\setlength{\tabcolsep}{5pt}
\resizebox{\textwidth}{!}{
\begin{tabular}{l l c c c c c}
\toprule
\multirow{2}{*}{\textbf{Dataset}} 
& \multirow{2}{*}{\textbf{Method}} 
& \multirow{2}{*}{\textbf{\# Samples}} 
& \textbf{Avg. Time /} 
& \textbf{Total Sample} 
& \textbf{Peak Alloc.} 
& \textbf{Peak Reserved} \\
& & & \textbf{Sample (s)} 
& \textbf{Time (s)} 
& \textbf{Mem. (GB)} 
& \textbf{Mem. (GB)} \\
\midrule
\multirow{5}{*}{\textbf{HRBench8K}}
& Qwen2.5-VL-7B (Base) & 800 & 6.8 & 5453.3 & 18.9 & 21.6 \\
\cmidrule(lr){2-7}
& \cellcolor{gray!20}\textbf{UniVLR (Ours)} 
& \cellcolor{gray!20}800 
& \cellcolor{gray!20}\textbf{4.5} {\scriptsize \textcolor{teal}{\textbf{$\downarrow$33.8\%}}}
& \cellcolor{gray!20}\textbf{3620.1} {\scriptsize \textcolor{teal}{\textbf{$\downarrow$33.6\%}}}
& \cellcolor{gray!20}\textbf{17.4} {\scriptsize \textcolor{teal}{\textbf{$\downarrow$7.9\%}}}
& \cellcolor{gray!20}\textbf{18.9} {\scriptsize \textcolor{teal}{\textbf{$\downarrow$12.5\%}}} \\
& CoVT & 800 & 9.4 & 7559.1 & 19.0 & 22.0 \\
& LVR & 800 & 5.0 & 4024.0 & 17.4 & 18.9 \\
& SkiLa & 800 & 9.4 & 7546.4 & 17.4 & 18.9 \\
\midrule
\multirow{5}{*}{\textbf{MME-RealWorld-Lite}}
& Qwen2.5-VL-7B (Base) & 1919 & 4.3 & 8251.7 & 19.0 & 21.7 \\
\cmidrule(lr){2-7}
& \cellcolor{gray!20}\textbf{UniVLR (Ours)} 
& \cellcolor{gray!20}1919 
& \cellcolor{gray!20}\textbf{3.2} {\scriptsize \textcolor{teal}{\textbf{$\downarrow$25.6\%}}}
& \cellcolor{gray!20}\textbf{6076.7} {\scriptsize \textcolor{teal}{\textbf{$\downarrow$26.4\%}}}
& \cellcolor{gray!20}\textbf{17.4} {\scriptsize \textcolor{teal}{\textbf{$\downarrow$8.4\%}}}
& \cellcolor{gray!20}\textbf{18.9} {\scriptsize \textcolor{teal}{\textbf{$\downarrow$12.9\%}}} \\
& CoVT & 1919 & 4.5 & 8664.1 & 19.0 & 22.0 \\
& LVR & 1919 & 3.6 & 6859.8 & 17.4 & 18.9 \\
& SkiLa & 1919 & 4.7 & 8966.3 & 17.5 & 18.9 \\
\bottomrule
\end{tabular}
}
\end{table}

Table~\ref{tab:additional_efficiency} shows that UniVLR provides favorable runtime efficiency under the actual evaluation setting. 
On HRBench8K, UniVLR reduces the average per-sample time from 6.8 seconds for Qwen2.5-VL-7B to 4.5 seconds, corresponding to a 33.8\% reduction. 
The total sample time is also reduced from 5453.3 seconds to 3620.1 seconds. 
Compared with prior visual latent reasoning methods, UniVLR is faster than LVR and substantially faster than CoVT and SkiLa, which require 9.4 and 9.4 seconds per sample, respectively.

A similar trend is observed on MME-RealWorld-Lite. 
UniVLR reduces the average per-sample time from 4.3 seconds for the base model to 3.2 seconds, and reduces the total sample time from 8251.7 seconds to 6076.7 seconds. 
Among visual latent reasoning baselines, UniVLR is also faster than CoVT, LVR, and SkiLa under the same evaluation protocol.

UniVLR further maintains a compact memory footprint. 
On HRBench8K, it reduces peak allocated memory from 18.9 GB to 17.4 GB and peak reserved memory from 21.6 GB to 18.9 GB compared with the base model. 
On MME-RealWorld-Lite, UniVLR similarly reduces peak allocated memory from 19.0 GB to 17.4 GB and peak reserved memory from 21.7 GB to 18.9 GB. 
Its memory usage is comparable to LVR and SkiLa, while remaining lower than CoVT on both benchmarks.

These results support the practical efficiency of the proposed unified visual latent reasoning interface. 
While the main paper focuses on reasoning-token efficiency, this additional evaluation shows that UniVLR does not introduce extra runtime or memory overhead in practice. 
Instead, the compact latent reasoning trajectory leads to lower latency and comparable or lower GPU memory usage than representative visual latent reasoning baselines. 
We emphasize that this experiment is intended as a lightweight efficiency check rather than a full systems-level throughput benchmark, since standardized throughput protocols for visual latent reasoning are not yet established and final textual outputs are often very short.


\end{document}